\documentclass{article}
\usepackage[utf8]{inputenc}
\usepackage{hyperref}
\usepackage{graphicx}
\usepackage{caption}
\usepackage{subcaption}
\usepackage{geometry}
\usepackage{amsmath}
\usepackage{amsfonts}
\usepackage{amssymb}
\usepackage{amsthm}
\usepackage{relsize}
\usepackage{nicefrac}
\usepackage{stackrel}
\usepackage[normalem]{ulem}
\usepackage{algorithm}
\usepackage{algpseudocode}
\usepackage{bbm}
\usepackage{bm}
\usepackage{fontawesome} %
\usepackage{bbding}    %
\usepackage{booktabs}

\usepackage{listings}
\usepackage{thmtools,thm-restate}

\newtheorem{claim}{Claim}%
\newtheorem{assumption}{Assumption}%

\theoremstyle{definition}

\theoremstyle{remark}

\def\env{\mathcal{E}}

\def\F{\mathbb{F}}

\def\Pr{\mathbb{P}}

\def\1{\mathbf{1}}

\newcommand{\ignore}[1]{}
\def\ind{\mathbbm{1}}

\def\sl{\mathrm{SL}}
\def\regavg{\mathrm{RegAvg}}

\usepackage{setspace}
\DeclareMathAlphabet{\mathdutchcal}{U}{dutchcal}{m}{n}

\usepackage[mathscr]{euscript}
\hypersetup{
    colorlinks=true,
    linkcolor=blue,
    filecolor=magenta,      
    urlcolor=purple,
    citecolor=blue,
}
\usepackage{xcolor}
\usepackage{tcolorbox}  %

\newcount\Comments
\newcommand{\kibitz}[2]{\ifnum\Comments=1{\textcolor{#1}{\textsf{\footnotesize #2}}}\fi}

\usepackage{natbib}
\title{Granular feedback merits sophisticated aggregation}

\usepackage{authblk}
\author[1]{Anmol Kagrecha}
\author[2]{Henrik Marklund}
\author[3]{Potsawee Manakul}
\author[4]{Richard Zeckhauser}
\author[1,5]{Benjamin Van Roy}

\affil[1]{Department of Electrical Engineering, Stanford University}
\affil[2]{Department of Computer Science, Stanford University}
\affil[3]{SCB 10X, SCBX Group}
\affil[4]{John F. Kennedy School of Government, Harvard University}
\affil[5]{Department of Management Science and Engineering, Stanford University}
\date{}

\begin{document}

\maketitle

\begin{abstract}
    Human feedback is increasingly used across diverse applications like training AI models, developing recommender systems, and measuring public opinion -- with granular feedback often being preferred over binary feedback for its greater informativeness. While it is easy to accurately estimate a population's distribution of feedback given feedback from a large number of individuals, cost constraints typically necessitate using smaller groups. A simple method to approximate the population distribution is regularized averaging: compute the empirical distribution and regularize it toward a prior. Can we do better? As we will discuss, the answer to this question depends on feedback granularity.

    Suppose one wants to predict a population’s distribution of feedback using feedback from a limited number of individuals. We show that, as feedback granularity increases, one can substantially improve upon predictions of regularized averaging by combining individuals' feedback in ways more sophisticated than regularized averaging.
    
    Our empirical analysis using questions on social attitudes confirms this pattern. In particular, with binary feedback, sophistication barely reduces the number of individuals required to attain a fixed level of performance. By contrast, with five-point feedback, sophisticated methods match the performance of regularized averaging with about half as many individuals.
\end{abstract}

\section{Introduction}
Human feedback plays a vital role in training AI models \citep{christiano2017deep, stiennon2020learning, nakano2021webgpt, ouyang2022training, bai2022training, touvron2023llama, grattafiori2024llama, liu2024deepseek, yang2025qwen3}, developing recommender systems \citep{ricci2010introduction, lu2012recommender}, and measuring social values \citep{gss_homepage, wvs_homepage, issp_homepage, pew_homepage}. Across these domains, granular feedback is consistently favored over binary feedback for its greater informativeness. In AI model training, researchers typically employ multi-point scales: \citet{christiano2017deep, stiennon2020learning, nakano2021webgpt, bai2022training, touvron2023llama, grattafiori2024llama} collect feedback on scales containing 3 to 19 points. Commercial recommendation platforms demonstrate similar practices, with \citet{primevideo_homepage} implementing 5-point rating systems and \citet{imdb_homepage} utilizing 10-point scales. Similarly, social science research organizations routinely employ multi-point scales, with major surveys like the General Social Survey, World Values Survey, International Social Survey Programme, and Pew Research Center commonly featuring 4- to 7-point scales. We refer to feedback on such scales as \textit{granular}, i.e., more differentiated than binary.

We demonstrate the greater informativeness of granular feedback using a simple example. Suppose a large population is asked to rate two public policies: one related to abortion and the other related to traffic management. On a binary scale, the distributions of the ratings might be equally skewed—perhaps 60\% approval for each policy. However, on a granular scale, the distribution of ratings for the abortion policy could be much more polarized than the ratings for the traffic management policy, with respondents clustering at the extremes rather than close to the center. Clearly, distributions on the granular scale are more useful for understanding the underlying population preferences. 

If the feedback reflects individuals' preferences, then collecting feedback from a large population can be valuable for developing systems and making decisions that accurately represent broader population preferences. However, seeking feedback from many individuals can be expensive. Therefore, it is common to work with a smaller group. A simple method to approximate the large population's feedback distribution is regularized averaging: compute the empirical distribution and regularize it towards a prior.

It is natural to ask if combining individuals' feedback in more sophisticated ways could lead to an improvement over predictions produced by regularized averaging. We show that the granularity of feedback significantly impacts the advantage of sophisticated aggregation methods, and this is our main contribution. Specifically, we provide empirical evidence and argue for the following claim: as feedback granularity increases, the advantage of sophisticated methods increases relative to regularized averaging. Moreover, the improvements can be significant even for modest granularity, such as 5-point scales. After discussing our main claim and the supporting evidence for it, we discuss how typical pipelines for training AI systems, such as large language models, could be improved by using sophisticated aggregation schemes (see Section~\ref{sec:rlhf}).

\begin{figure}[!htp]
    \centering
    \includegraphics[width=0.4\linewidth]{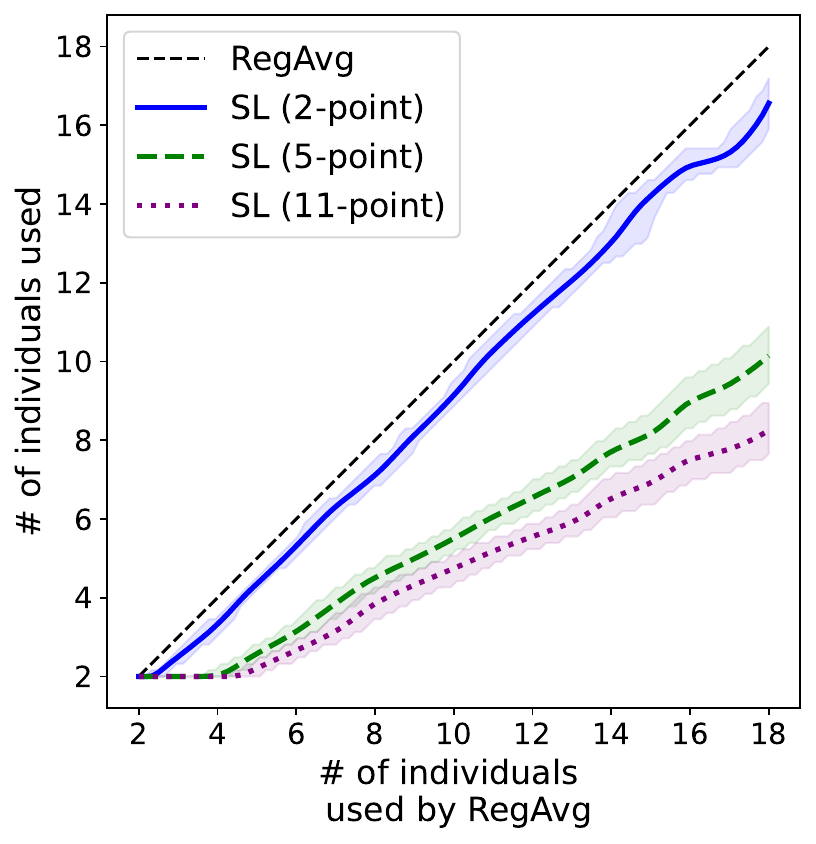}
    \caption{Performance comparison of $\sl$ and $\regavg$ for different granularities of feedback. Note that a smaller slope indicates better performance.}
    \label{fig:punchline_plot}
\end{figure}

Here are the key elements and results of our empirical study; further detail is provided in Section~\ref{sec:all_evidence}. In this study, we compare regularized averaging ($\regavg$) and supervised learning ($\sl$) on a dataset with questions about social attitudes and values. Moreover, this dataset has annotations from about forty individuals on all task units. We divide the individuals into two sets of roughly equal size: an input-set and an output-set. We let the input to the methods, $\regavg$ and $\sl$, come from individuals in the input-set. Moreover, we use the empirical distribution of the feedback from the individuals in the output-set to evaluate the methods' predictions. Figure~\ref{fig:punchline_plot} demonstrates our key finding: as granularity increases, the advantage of $\sl$ over $\regavg$ increases. With binary feedback, $\sl$ and $\regavg$ perform almost similarly. However, with 5-point and 11-point feedback, to match $\regavg$'s performance, $\sl$ requires about 44\% and 54\% fewer individuals, respectively.

Here is an overview of the argument for why our claim plausibly holds more generally. Further details are given in Section~\ref{sec:intuition}. Suppose we are trying to predict an element of the population's cumulative distribution function (CDF). We first show that, irrespective of the feedback granularity, regularized averaging uses a specific binarized form of each individual's feedback to predict this element. Thus, except when feedback itself is merely binary, it throws away useful information. Then, we argue that the discarded information typically increases with granularity. Finally, we argue that under certain conditions, methods like $\sl$ exploit all available information in the input about the quantity they are being trained to predict. Hence, if one uses $\sl$, or some qualitatively similar sophisticated method, it will likely have an advantage over $\regavg$ for granular data, and a somewhat greater advantage as granularity grows. 

\section{Related work}
\label{sec:rel_work}

\textbf{Overview.} Our work investigates when sophisticated aggregation methods could provide substantial benefits over simple methods like regularized averaging for predicting population distributions. To contextualize this investigation, we organize our review as follows. We first discuss the relationship between our distribution prediction problem and the extensive literature on ground truth estimation in crowdsourcing, noting that many existing aggregation methods could potentially be adapted for our setting. We then motivate why distribution prediction is particularly valuable in domains characterized by subjectivity, i.e., in domains where individuals may disagree about what constitutes the correct answer. Moreover, as granular feedback forms a key part of our investigation, we review evidence demonstrating that granular feedback improves performance across diverse applications. We then discuss methods such as the Delphi technique that allow individuals to update their feedback and provide some reasons for our focus on a problem setting where individuals provide feedback only once.

\textbf{Aggregation algorithms for crowdsourced feedback.} A large body of work in crowdsourcing focuses on aggregating individual feedback to estimate ground truth labels, where the goal is to recover a single correct answer for each task unit. Most work has focused on problems where the feedback from crowdworkers is unordered \citep{zhang2016learning, zheng2017truth, zhang2022knowledge}, though some works propose methods to aggregate ordered or granular feedback \citep{lakshminarayanan2013inferring, zhou2014aggregating, metrikov2015aggregation, guo2016aggregating, nguyen2016probabilistic, chen2017learning, li2020gpm}. These methods assume that an underlying true label exists and that individual responses contain noise around this truth. However, many of these aggregation algorithms could potentially be repurposed for predicting population distributions, even though they were originally designed for ground truth estimation. Our work addresses a different problem: predicting population distribution when individuals' feedback is ordered. We investigate when sophisticated aggregation methods provide substantial benefits over simple methods in this setting, rather than proposing new aggregation algorithms.

\textbf{Subjectivity and the value of distribution prediction.} In many domains, individuals exhibit disagreement about what constitutes the correct answer, suggesting that ground truth may not exist or may be poorly defined. For example, people disagree about toxicity detection \citep{gordon2021disagreement}, ethical judgments \citep{lourie2021scruples}, and inferring entailment \citep{pavlick2019inherent}. In these domains, the empirical distribution of feedback does not concentrate on a single value despite collecting feedback from many individuals. Given this subjective nature of individuals' feedback, predicting the full population distribution of responses becomes a natural and valuable approach \citep{zhang2017predicting, liu2019learning, wang2019classification}. Motivated by the value of distribution prediction, we focus on this problem setting in our work.

\textbf{Works demonstrating usefulness of granular feedback.} Across multiple domains, research has demonstrated that granular feedback scales lead to improved performance compared to binary scales. In reinforcement learning with human feedback, \citet{holladay2016active} show that 3-point feedback requires fewer queries to humans to reach the same performance compared to binary feedback, while \citet{wilde2021learning} observe that quasi-continuous feedback leads to more accurate reward function estimation than binary feedback. Similarly, \citet{yu2023thumbs} demonstrate that 11-point feedback achieves higher rewards than binary feedback baselines. In forecasting, \citet{friedman2018value} find that crowdworkers produce more accurate forecasts when allowed to use more granular scales rather than binary predictions. For clinical trials, \citet{mchugh2010simulation}, \citet{roozenbeek2011added}, and \citet{d2020ordinal} show that granular outcome scales make it easier to differentiate between treatments compared to binary outcomes. These findings establish that granular feedback provides benefits across diverse applications. Given this evidence, population distributions arising from granular feedback scales could be more informative than those from binary scales, motivating our investigation into how feedback granularity affects the performance of different aggregation methods for distribution prediction.

\textbf{Iterative feedback methods and our focus on single-round aggregation.} When seeking to predict or estimate quantities that are not easily knowable, it is common to collect estimates from multiple individuals and aggregate them. The Delphi method \citep{dalkey1963experimental, linstone1975delphi} is a foundational and influential approach for aggregating estimates, where participants provide estimates across multiple rounds and receive some information about others' feedback between rounds to refine their estimates. Extensions of this approach include real-time Delphi systems, which allow continuous updating \citep{gnatzy2011validating}, and prediction markets where participants can continuously revise their estimates based on market prices \citep{wolfers2004prediction}. In forecasting tournaments, participants regularly update their estimates as new information becomes available \citep{ungar2012good, baron2014two, tetlock2016superforecasting}. While these iterative methods can be highly effective, they require multiple rounds of interaction and may not be feasible in all settings. For analytical tractability and to focus on the core question of how feedback granularity affects aggregation performance, we restrict our investigation to a problem setting where individuals provide feedback for a task unit only once.

\section{Problem formulation}
\label{sec:prob_formulation}
Consider an infinite population of individuals. In round $t$, these individuals observe a task unit $U_t$. An example of a task unit is a tuple containing a prompt, a response, and a text asking to rate the helpfulness of the response. After observing this unit, the individuals give feedback $(Y_{t,k}: k \in \mathbb{N})$. For any $k$, preference $Y_{t,k}$ lies in a space $\mathcal{Y} \subset [-1,1]$. Moreover, we assume that $|\mathcal{Y}|<\infty$. 

We will make things precise next, but at a high level, the problem is to predict the distribution of the feedback sequence $(Y_{t,k}: k \in \mathbb{N})$ using feedback from finitely many individuals. We call the entity that needs to make predictions the \textit{center}. To model uncertainty from the center's perspective, we take the preference sequence $(Y_{t,k}: k \in \mathbb{N})$ to be a random variable. This sequence and all other random variables we consider are defined with respect to a probability space $(\Omega, \F, \Pr)$. We further make the assumption that the sequence $(Y_{t,k}: k \in \mathbb{N})$ is exchangeable. For a concise introduction to exchangeability, see Section~2 of the work by \citet{barber2024finetti}. A consequence of this exchangeability assumption is that for all $y \in \mathcal{Y}$, $\lim_{\Tilde{K} \to \infty} \frac{\sum_{k=1}^{\Tilde{K}} \ind[Y_{t,k} \leq y]}{\Tilde{K}}$ exists almost surely. We call this collection of limits the population CDF and denote it by $P_t$, i.e.,
\begin{align}
    \label{eq:pop_cdf}
    P_t(y) = \lim_{\Tilde{K} \to \infty} \frac{\sum_{k=1}^{\Tilde{K}} \ind[Y_{t,k} \leq y]}{\Tilde{K}}.
\end{align}
The center observes feedback from the first $K$ individuals in each round $t$. Moreover, it can use all feedback $Y_{1:t,1:K} \in \mathcal{Y}^{t \times K}$ observed till the current round and past population CDFs $P_{1:t-1}$ to produce a predictive CDF $Q_t$.

We now introduce the notion of a \textit{policy}. We call a function from the sequence $(Y_{1,1:K}, P_1, \cdots, \allowbreak Y_{t-1,1:K}, P_{t-1}, Y_{t,1:K})$ to $Q_t$ as a policy. As the center processes this sequence for each round $t$ and produces a predictive CDF, we can think of the center as executing a policy. Moreover, when comparing policies $\pi$ and $\pi'$, we will use $Q_t^{(\pi)}$ and $Q_t^{(\pi')}$ to denote their respective predictions, but in other cases, we will skip the superscript.

We do not use the standard log loss for evaluation, but rather use a variant. We first discuss why we don't use the standard log loss and then discuss the variant. Loosely speaking, the reason for not using the standard log loss is that the elements of the feedback space $\mathcal{Y}$ can be ordered. For example, the elements of $\mathcal{Y}$ can be -1, 0, and 1, but not cat, dog, and bunny. Since the elements can be ordered, mistaking adjacent elements should be penalized less than mistaking faraway elements. Using an example, we demonstrate that this is not the case when one uses the standard log loss. Suppose $\mathcal{Y}=\{-1,0,1\}$. Let $p_t$ be the probability mass function (PMF) corresponding to CDF $P_t$. Suppose the values taken by $p_t$ at $-1$, $0$, and $1$ are $(0,0,1)$, respectively. For notational convenience, we will just say $p_t=(0,0,1)$ and index the elements of $p_t$ by -1, 0, and 1. Now, consider two predictive CDFs $Q_t$ and $Q_t'$. Let the corresponding PMFs be $q_t=(0.005, 0.99, 0.005)$ and $q_t'=(0.99, 0.005, 0.005)$. As $p_{t,1}=1$, $q_{t,0} \approx 1$ and $q_{t,-1}' \approx 1$, we want $q_t$ to perform better than $q_t'$. However, as shown by calculations in Table~\ref{tab:loss_motiv}, this is not the case.

Here is the variant of the standard log loss we use:
\begin{equation}
\label{eq:loss_fn}
\mathcal{L}(P_t, Q_t)= - \frac{1}{|\mathcal{Y}|-1} \sum_{y \in \mathcal{Y}} \left(P_t(y) \log Q_t(y) + \left(1-P_t(y)\right) \log \left(1-Q_t(y)\right)\right).
\end{equation}
While standard log loss evaluates the probabilities of different scores, this loss function applies log loss to cumulative probabilities. Moreover, the calculations in Table~\ref{tab:loss_motiv} demonstrate how this loss function takes the ordinal nature of the feedback into account. Additionally, this loss function is often applied to classification problems with ordinal labels \citep{niu2016ordinal, cao2020rank, shi2023deep}.

\begin{table}[htp]
    \centering
    \begin{tabular}{|c|c|c|c|}
    \toprule
    \textbf{PMF} & \textbf{log loss} & \textbf{CDF} & \textbf{Loss in} \textbf{Eqn.}~\ref{eq:loss_fn} \\
    \midrule
    
    $p_t=(0, 0, 1)$ & 0 & $P_t=(0,0,1)$ & $0$ \\[2pt]
    
    $q_t=(0.005,0.99,0.005)$ & $-\log 0.005$ & $Q_t=(0.005,0.995,1)$ & $-\frac{1}{2} \left( \log 0.995 + \log 0.005 \right)$ \\[2pt]
    
    $q_t'=(0.99,0.005,0.005)$ & $-\log 0.005$ & $Q_t'=(0.99,0.995,1)$ & $-\frac{1}{2} \left( \log 0.01 + \log 0.005 \right)$ \\[2pt]
    \bottomrule
    \end{tabular} \\[3pt]
    \caption{Standard log loss does not respect the ordinal nature of the feedback but loss function \eqref{eq:loss_fn} does. PMFs $q_t$ and $q_t'$ are predictions for $p_t$. As argued in the text, we want $q_t$ to be penalized less than $q_t'$. However, the standard log loss is the same for both predictions. One can check that the calculations for loss function \eqref{eq:loss_fn} are correct but here is some intuition for why the values end up being what they are. Observe that $|Q_t'(y)-P_t(y)|$ is large for $y \in \{-1,0\}$ while $|Q_t(y)-P_t(y)|$ is large only at $y=0$. Because of these errors and the fact that the loss function \eqref{eq:loss_fn} is in terms of CDFs and not PMFs, $q_t'$ ends up having a larger loss than $q_t$.}
    \label{tab:loss_motiv}
\end{table}

\section{Dataset on social values and attitudes}
\label{sec:dataset}
We have collected a new dataset and will perform experiments using it. The details about the experiments are provided in the next section. Here, we provide the rationale for collecting this dataset and outline its main features. This dataset is available \href{https://huggingface.co/datasets/scb10x/survey_social_value_th2025}{here}.

\subsection{Reasons for collecting a new dataset}
\label{sec:reason_new_dataset}
Our decision to collect a new dataset was primarily influenced by the kind of analysis we wanted to perform. We were interested in investigating the following. One, the performance of different policies, as there is an increase in the number of individuals giving feedback. Two, the change in their performance as we increase feedback granularity. Given the need to perform such analysis, we required a large enough number of individuals and feedback granularity. Roughly speaking, these requirements were missing from popular RLHF datasets \citep{awesome_llm_human_preference_datasets} and datasets in the crowdsourcing literature like those by \citet{AdultContent3-HCOMP2010}, \citet{TREC-RF10-Crowd},  \citet{zhou2012learning}, and \citet{halpern2016mobile}. We give more details in Appendix~\ref{app:reason_new_dataset}.

\subsection{Main features of the dataset we collected}

Each task unit contains a question related to social attitudes and values and was generated using an LLM. One sample task unit is given in Figure~\ref{pg:example_task_units} and more are mentioned in Appendix~\ref{app:ques_gen}. Moreover, the full dataset is available with the supplementary material. We constructed such task units because they can be quickly answered and because we found the topic of social attitudes and values interesting. The total number of task units in this dataset is 1020, and we get 11-point feedback from thirty-nine crowdworkers on all task units. The only demographic information we know about the crowdworkers is that they were from Thailand and were proficient in the Thai language. We were unaware of how proficient they were in English, so we gave them Thai translations of the task units we designed. More details on how we generated these task units and collected the feedback are available in Appendix~\ref{app:dataset_creation}.

\newtcolorbox{contextbox}{
    colback=blue!10, %
    colframe=blue!60, %
    boxrule=0.5mm,  %
    width=\textwidth,
    sharp corners
}

\begin{figure}[htp]
\begin{minipage}{\textwidth}
\centering
\begin{contextbox}
\textbf{Context.}
Political discussions often happen at family gatherings like Songkran and New Year celebrations.

\textbf{Question.} How comfortable do you feel discussing politics with family members of different generations during holiday gatherings?

\textbf{Score.} -5 = Very uncomfortable, 0 = Neutral, +5 = Very comfortable
\end{contextbox}

\end{minipage}
\caption{An example of a task unit that we gave to the crowdworkers.}
\label{pg:example_task_units}
\end{figure}

\section{Experimental evidence and an argument supporting our main claim}
\label{sec:all_evidence}
Recall our main claim: as feedback granularity increases, there is an increase in the advantage of sophisticated policies over regularized averaging. In this section, we provide experimental evidence and an argument supporting this claim. We begin by explaining how the dataset on social values and attitudes is processed for our experiments. After that, we give details about the policies we consider and compare their performance for feedback of different granularities. Finally, we argue why our claim would hold beyond the current experimental setup.

\subsection{Data preparation}
\label{sec:data_prep}

We want to evaluate policies based on how well they can predict the population distribution of feedback. This kind of evaluation is ideal but impractical because the population distribution is the distribution of feedback from an infinite number of individuals. To approximate this idealized evaluation, we divide individuals into two sets: an input-set and an output-set. Here, we did this division using NumPy's random number generator; more details in Appendix~\ref{app:data_prep_details}. As the names of these sets suggest, the feedback from the input-set individuals forms the input to policies, and the feedback from the output-set individuals is used to evaluate the predictions of policies. Moreover, the number of input-set individuals and output-set individuals are roughly equal: twenty and nineteen, respectively. We denote these numbers as $L$ and $L'$, respectively. Additionally, we denote the feedback given by output-set individuals for task unit $U_t$ as $Y'_t \in \mathcal{Y}^{L'}$ and the empirical CDF of output-set individuals as $O_t$. Formally, for all preference scores $y \in \mathcal{Y}$,
\begin{align*}
    O_t(y) := \frac{\sum_{l=1}^{L'} \ind[Y'_{t,l} \leq y]}{L'}.
\end{align*}
For ease of notation, we call $O_t$ the \textit{output-set CDF}.

We want to vary the number of individuals $K$ and compute the average loss across all subsets of the input-set of size $K$. This is too expensive. A cheaper alternative is to sample many subsets. Instead of independently sampling random subsets for each $K$, we use a nested sampling approach so that we can check how loss changes as we add feedback from more individuals. Specifically, we sample multiple random permutations of the sequence $(1,\cdots,L)$ and use the feedback from the first $K$ individuals as input to a policy. We refer to such a permutation as an \textit{environment}. We used thirty environments for hyperparameter tuning and thirty for evaluation. For ease of notation, we will call these sets of environments \textit{train-environments} and \textit{eval-environments} respectively.

We did not collect a separate dataset with binary or 5-point feedback. To create the binary dataset, we replaced negative scores by -1, positive scores by +1, and randomly assigned neutral scores to $\{-1,1\}$. For the 5-point dataset, we bin the 11-point scores into five categories (see Appendix~\ref{app:data_prep_details} for details).

\subsection{Policies}
\label{sec:policies}
We now describe two policies that we will evaluate. First, regularized averaging. We consider this, given its simplicity and ubiquity. Matching its performance is a bare minimum for any sophisticated policy. The second policy that we consider is supervised learning. We choose supervised learning because it has enjoyed great success in many applications and could be beneficial here.

\textbf{Regularized averaging} or $\regavg$ first computes the empirical distribution of input feedback and regularizes it towards a prior distribution. More formally, for an environment $\env$, the predictive CDF $Q_t$ of $\regavg$ is:
\begin{align}
    \label{eq:regavg_pred}
    Q_t(y) = \gamma Q_0(y) + (1-\gamma) \frac{\sum_{k=1}^K \ind[Y_{t,\env[k]} \leq y]}{K},
\end{align}
where $\gamma$ is a hyperparameter and $Q_0$ is some initial CDF. In our experiments, we let $Q_0$ be the average of the empirical CDFs of OOS individuals: $Q_0(y) = \frac{\sum_{\tau \leq T} O_t(y)}{T}$, where $T$ is the size of our dataset. Having chosen $Q_0$, we tuned $\gamma$. For each $K$, we varied $\gamma$ in $(0.1, 0.2, \cdots, 1.0)$ and chose the value that led to the best performance on the training environments.

\textbf{Supervised learning} or $\sl$ learns to predict the empirical distribution of output-set individuals using feedback from a subset of input-set individuals. Formally, for an environment $\env$ and number of individuals $K$, $\sl$ learns to predict $O_\tau$ from $Y_{\tau, \env[1:K]}$. Let $\mathcal{D}_{\env,K} = \{ (Y_{\tau, \env[1:K]}, O_\tau): \tau \leq T \}$. We perform 5-fold cross validation on $\mathcal{D}_{\env,K}$. Our supervised learning model is a slightly modified version of a standard multi-layer perceptron (MLP), and this ensures that the output of the MLP is a valid CDF. We train this MLP using standard deep learning regularization and optimization techniques. More details about this architecture, the training process, and the amount of computing resources used are given in Appendix~\ref{app:sl_hyperparam}.

\subsection{Empirical evidence for our main claim}
\label{sec:exp_evidence}
Here, we give empirical evidence for our main claim that the advantage of sophisticated policies over regularized averaging increases as feedback becomes more granular. Specifically, we do so by comparing the performance of $\regavg$ and $\sl$ for 2-point, 5-point, and 11-point feedback. 

\begin{figure}[htp]
    \centering
    \begin{subfigure}{0.3\textwidth}
        \centering
        \includegraphics[width=\textwidth]{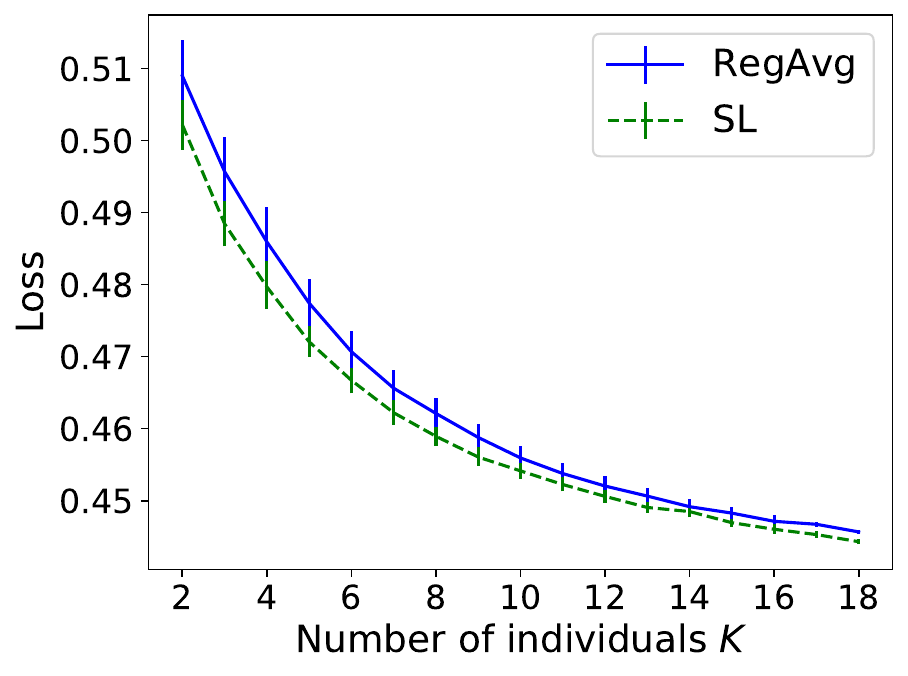}
        \caption{2-point feedback}
        \label{fig:loss_vs_K_2}
    \end{subfigure}
    \hfill
    \begin{subfigure}{0.3\textwidth}
        \centering
        \includegraphics[width=\textwidth]{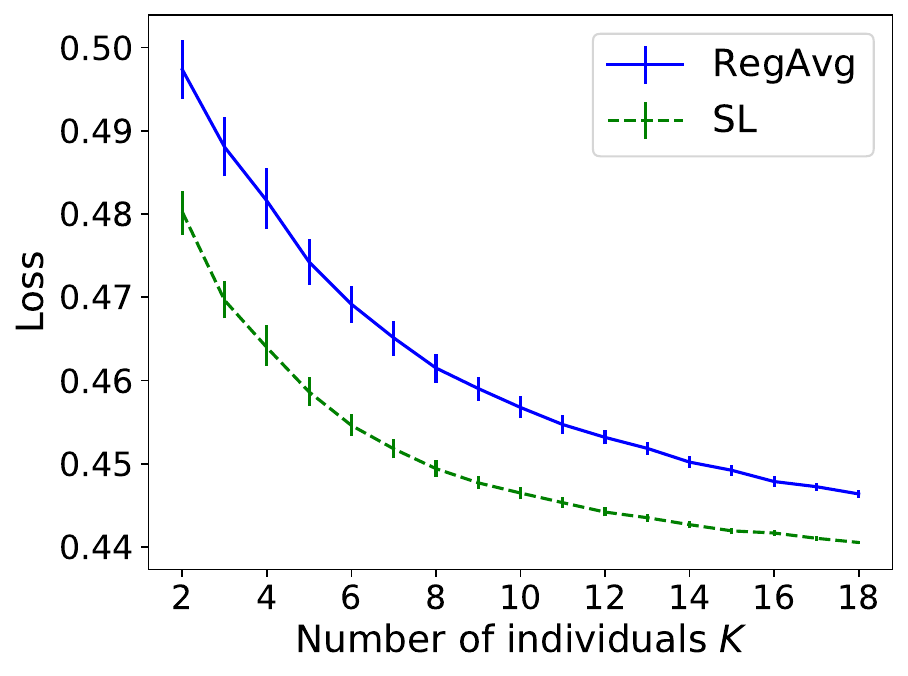}
        \caption{5-point feedback} 
        \label{fig:loss_vs_K_5}
    \end{subfigure} 
    \hfill
    \begin{subfigure}{0.3\textwidth}
        \centering
        \includegraphics[width=\textwidth]{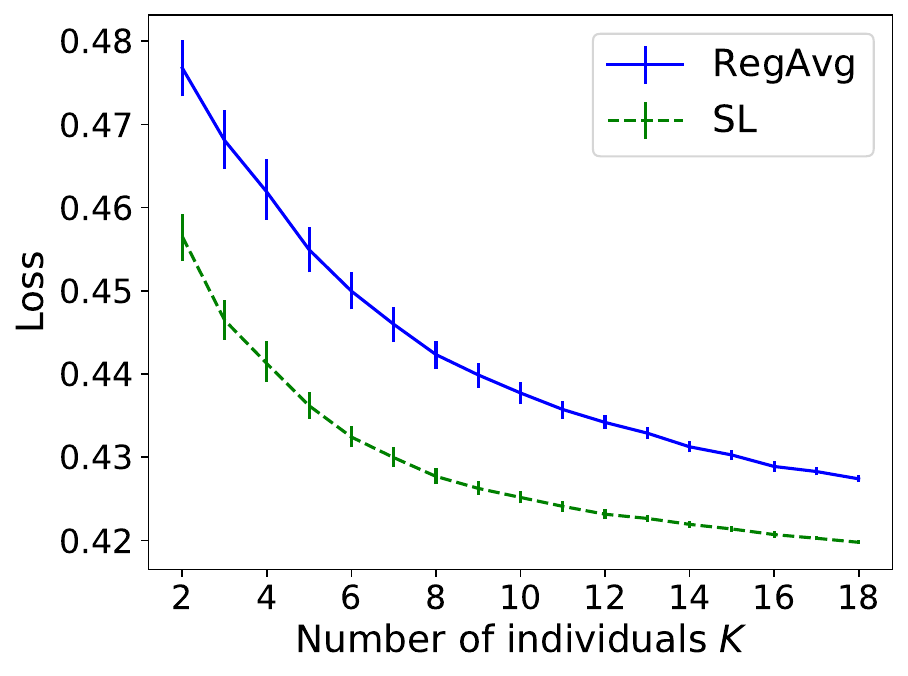}
        \caption{11-point feedback} 
        \label{fig:loss_vs_K_11}
    \end{subfigure}
    \caption{Losses of policies as a function of number of individuals for feedback of different granularities.}
    \label{fig:loss_vs_K}
\end{figure}

In Figure~\ref{fig:loss_vs_K}, we report the mean loss incurred by the policies for feedback of different granularities and with different numbers of individuals. To compute the mean loss, we average over eval-environments and task units; see Section~\ref{sec:data_prep} for the definition of the eval-environments. As mentioned in Section~\ref{sec:policies}, the evaluation in any environment is done using 5-fold cross-validation. Moreover, the factor of variability for the confidence intervals in Figure~\ref{fig:loss_vs_K} is the environment. We use bootstrap to compute the confidence interval at 95\% confidence, and we do so by generating 1000 bootstrap samples.

\begin{figure}
    \centering
    \includegraphics[width=0.4\linewidth]{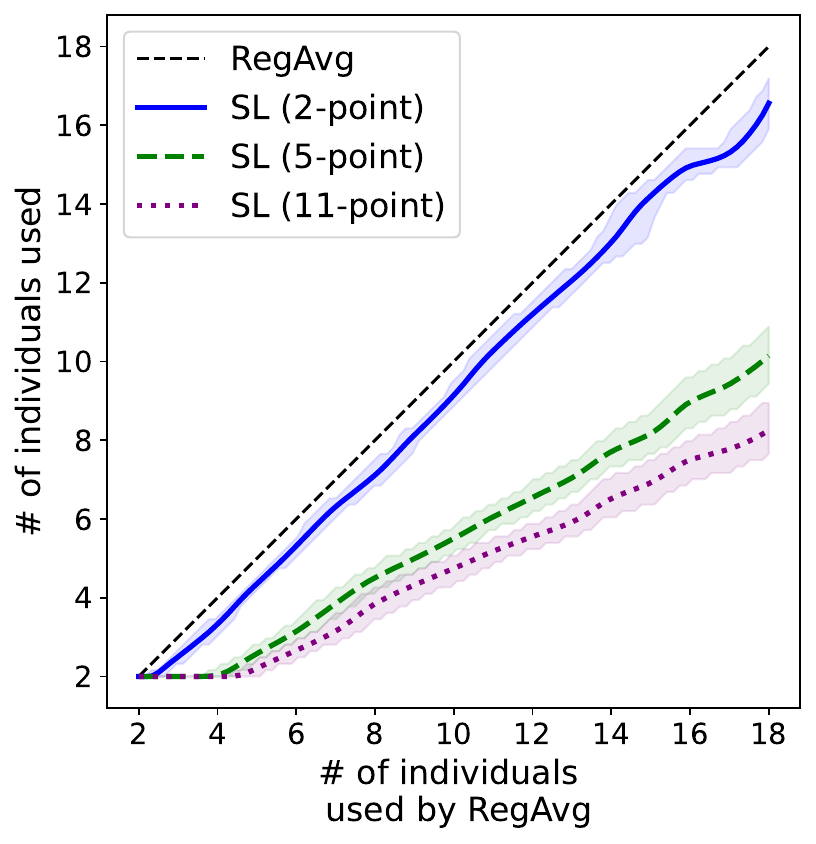}
    \caption{Number of individuals needed to match $\regavg$'s performance. Note here that a smaller slope means better performance. When feedback is on a 5-point scale, $\sl$ requires 44\% fewer individuals to achieve a performance equivalent to $\regavg$. For example, when $\regavg$ uses eighteen individuals, $\sl$ requires only about ten individuals. On the contrary, for 2-point feedback, $\sl$ requires about seventeen individuals to match the performance of $\regavg$ using eighteen individuals.}
    \label{fig:num_workers_comp}
\end{figure}

Raw loss values are often hard to interpret, so we compare the number of individuals used by policies to achieve similar performance; see Figure~\ref{fig:num_workers_comp}. Specifically, we vary the number of individuals used by $\regavg$ and find the number of individuals required by $\sl$ to match the performance of $\regavg$. To get a continuous plot, we interpolate between known mean losses at integer number of individuals. Both policies have known mean losses for integer values between 2 and 18. We use these integer values as the basis for interpolation and this allows us to estimate performance for fractional values. For Figure~\ref{fig:num_workers_comp}, we use a uniform grid with a hundred values starting at 2 and ending at 18.

Similar to Figure~\ref{fig:loss_vs_K}, the factor of variability for the confidence intervals for Figure~\ref{fig:num_workers_comp} is the environment. We use bootstrap to compute the confidence intervals at 95\% confidence. Specifically, we resample the set of environments 1000 times. For each set of environments, we potentially get different mean loss curves for both policies. These different mean loss curves lead to different requirements on the number of individuals to reach a certain performance, and this is what we plot in Figure~\ref{fig:num_workers_comp}.

\subsection{Argument for why our main claim would extend beyond current experimental setup}
\label{sec:intuition}

Our main claim is that the advantage of sophisticated policies relative to regularized averaging increases as feedback becomes more granular, and we provide an argument for why this claim would hold even for task units different from the ones for which we conducted our experiments. While the entire argument can be made mathematically formal, we choose not to do so because the main ideas can be communicated much more quickly when we keep our argument slightly informal.

We know that the outputs of $\regavg$ are predictions for the output-set CDF $O_t$, and our first claim below characterizes the predictions of $\regavg$ for an output-set CDF element $O_t(y)$ for any $y$. Before we state the claim, without loss of generality, we assume that all individuals' feedback has been scaled to be in $[-1,1]$.

\begin{claim}
    \label{claim:regavg_binary}
    When predicting output-set CDF $O_t(y)$ for any $y$, $\regavg$ only uses binarized feedback.
\end{claim}
\noindent The following is the justification. By definition, the prediction of $O_t(y)$ by $\regavg$ is
\begin{align*}
    Q_t(y) = \gamma Q_0(y) + (1-\gamma) \frac{1}{K} \sum_{k=1}^K \underbrace{\ind[Y_{t,k} \leq y]}_{{\rm binary\ variable}},
\end{align*}
where $\gamma$ is a hyperparameter and $Q_0$ is some prior distribution. We can observe that $\regavg$ is binarizing $Y_{t,k}$ to produce its prediction. For ease of notation, let $B_t^{(y)}=(\ind[Y_{t,k} \leq y]: k=1,\cdots,K)$ be the binarized feedback that $\regavg$ uses.

Next, we state an assumption about how much information the binarized feedback $B_t^{(y)}$ and the original feedback $Y_t$ provide about the output-set CDF element $O_t(y)$.

\begin{assumption}
\label{asmp:info_diff_increases}
    Fix a score $y \in [-1,1)$. Consider the difference in the information that one gets from $Y_t$ about $O_t(y)$ and the information one gets from $B_t^{(y)}$ about $O_t(y)$. As feedback granularity increases, this difference increases.
\end{assumption}
\noindent The following is a justification for this assumption. The more granular the feedback we have, the larger the information loss when we binarize. For example, when we just have binary feedback, we can recover feedback $Y_{t}$ from $B_{t}^{(y)}$. But, for more granular feedback, this is not the case because multiple scores are either greater or smaller than $y$, and depending on whether they are greater or smaller, they get mapped to 1 and 0, respectively. Moreover, knowing the precise scores a individual gives can be informative about the output-set CDF. For example, suppose a individual is harsher than others in the sense that they typically give more negative scores than others. Knowing how negative their score is informs how positive other individuals' scores will be, which, in turn, informs how positively skewed the output-set CDF will be. When we binarize, we lose information about how negative the score was, and thereby, lose information relevant to the output-set CDF. Given that binarizing leads to more information loss for more granular feedback and that some of the lost information was relevant to the output-set CDF, we can infer that binarizing more granular feedback could lead to more information loss about the output-set CDF. Given this inference, we make the assumption above.  

Claim~\ref{claim:regavg_binary} and Assumption~\ref{asmp:info_diff_increases} are the main components of the argument. If Assumption~\ref{asmp:info_diff_increases} holds, then, as granularity increases, sophisticated methods like $\sl$, that use the original feedback, have increasingly more information compared to $\regavg$ and can increasingly perform better. The rest of the argument provides further details on how this occurs.

Let $\sl_b$ be another supervised learning policy. The input to this policy is the original feedback $Y_t$, but when predicting an output-set CDF element $O_t(y)$, it only uses $B_t^{(y)}$. We next make an assumption about how $\sl$ and $\sl_b$ use the information available to them to predict $O_t(y)$.

\begin{assumption}
    \label{asmp:optimal_info_use}
    $\sl_b$ optimally uses all information in $B_t^{(y)}$ relevant to predicting $O_t(y)$ and $\sl$ optimally uses all information in $Y_t$ relevant to predicting $O_t(y)$.
\end{assumption}
\noindent If we have a large dataset and use large enough neural networks for $\sl_b$ and $\sl$, then this assumption will likely hold. 

\textbf{Definition of advantage.} We have used the word advantage loosely till now, but we will define it more precisely for the subsequent discussion. Let $\pi$ be a policy that incurs a mean loss $\ell$ when using $K$ individuals. Suppose that $\pi'$ needs to use at least $K'$ individuals to ensure that it incurs a mean loss less than $\ell$. Then at $K$, the advantage of $\pi'$ over $\pi$ is $\frac{K}{K'}$.

Now, we state our main claim more formally with $\sl$ as a candidate for a sophisticated policy. 
\begin{claim}
    \label{claim:adv_full_loss}
    With the assumptions above, for any $K$, the advantage of $\sl$ over $\regavg$ increases with feedback granularity.
\end{claim}
\noindent To show Claim~\ref{claim:adv_full_loss}, we use the structure of the loss function in Equation~\ref{eq:loss_fn}. Specifically, in this loss function, we average the log-losses in predicting $O_t(y)$ for different $y$. Given this, we argue that it is sufficient to show the following claim.

\begin{claim}
    \label{claim:adv_loss_element}
    With the assumptions above, for any $K$ and $y \in [-1,1)$, the advantage of $\sl$ over $\regavg$ in predicting $O_t(y)$ increases with feedback granularity.
\end{claim}
Here is the reason why Claim~\ref{claim:adv_full_loss} follows from Claim~\ref{claim:adv_loss_element}. First, we can ignore the case where $y=1$ because $O_t(1)$ is always one and both policies incur zero loss in predicting this element. Now, consider two feedback granularities $g_1<g_2$. Then, from Claim~\ref{claim:adv_loss_element}, for each $y$ and $K$, the advantage of $\sl$ over $\regavg$ in predicting $O_t(y)$ is larger for granularity $g_2$ than $g_1$. A larger advantage implies that the loss in predicting $O_t(y)$ decreases more quickly with the number of individuals. As our loss function~\ref{eq:loss_fn} for predicting $O_t$ averages losses at different values of $y$, the overall loss function will also decrease more quickly with $K$. This is equivalent to Claim~\ref{claim:adv_loss_element}.

To show Claim~\ref{claim:adv_loss_element}, we make three claims next, show these hold under the assumptions above, and then combine them to show Claim~\ref{claim:adv_loss_element}. 

\begin{claim}
    \label{claim:sl_b_better_than_regavg}
    With the assumptions above, for any feedback granularity, $\sl_b$ incurs a smaller loss than $\regavg$ in predicting $O_t(y)$ for each $K$.
\end{claim}
\noindent Claim~\ref{claim:sl_b_better_than_regavg} follows from Assumption~\ref{asmp:optimal_info_use}, i.e., $\sl_b$ optimally uses all information in $B_t^{(y)}$ relevant to predicting $O_t(y)$ while $\regavg$ may not. 

\begin{claim}
    \label{claim:monotonicity}
     With the assumptions above, for any feedback granularity, losses of $\sl$ and $\sl_b$ in predicting $O_t(y)$ decrease with $K$.
\end{claim}
\noindent As the number of individuals increases, $\sl$ and $\sl_b$ have feedback from one extra individual. Hence, the inputs to both are potentially more informative about $O_t(y)$. Then, by Assumption~\ref{asmp:optimal_info_use}, the losses for $\sl$ and $\sl_b$ will be smaller. 

\begin{claim}
   \label{claim:difference_increase} 
    Consider the difference in the losses incurred by $\sl_b$ and $\sl$ in predicting $O_t(y)$. With the assumptions above, for any $K$, as feedback granularity increases, this difference increases.
\end{claim}
\noindent Claim~\ref{claim:difference_increase} follows from Assumptions \ref{asmp:info_diff_increases} and \ref{asmp:optimal_info_use}. Here's why. Loosely speaking, Assumption~\ref{asmp:info_diff_increases} says that as granularity increases, the informativeness of $Y_t$ relative to $B_t^{(y)}$ about $O_t(y)$ increases. By Assumption~\ref{asmp:optimal_info_use}, we have that this increase in information in $Y_t$ is exploited by $\sl$. Hence, in predicting $O_t(y)$, the difference in losses between $\sl_b$ and $\sl$ increases with granularity.

Here is how Claim~\ref{claim:adv_loss_element} follows from Claims~\ref{claim:sl_b_better_than_regavg}-\ref{claim:difference_increase}. Using Claims~\ref{claim:monotonicity} and \ref{claim:difference_increase}, we have that for any score $y$ and $K$, there is an increase in the advantage of $\sl$ over $\sl_b$ in predicting $O_t(y)$ as granularity increases. Moreover, Claim~\ref{claim:sl_b_better_than_regavg} tells us that $\sl_b$ is as good as $\regavg$. This implies that the advantage of $\sl$ relative to $\regavg$ is lower bounded by the advantage of $\sl$ relative to $\sl_b$. As this lower bound increases with granularity, the advantage relative to $\regavg$ must increase.

\section{Sophisticated aggregation methods could improve data efficiency of typical RLHF pipelines}
\label{sec:rlhf}

Reinforcement learning with human feedback (RLHF) is a common component of training state-of-the-art large language models or LLMs \citep{achiam2023gpt, yang2025qwen3, team2024gemini, grattafiori2024llama, liu2024deepseek}. As suggested by its name, human feedback is a crucial component of this methodology. Such feedback enables updating of an LLM's parameters so that the LLM produces responses that are more strongly favored by humans. For more details, see survey articles on RLHF by \citet{casper2023open, kaufmann2023survey}. We will give an overview of the RLHF pipeline for the purpose of justifying the following claim. Consider LLMs produced using typical RLHF pipelines. We can match their quality at a fraction of the cost by following two steps: (1) perform small tweaks to feedback collection procedures in RLHF, and (2) use sophisticated methods to aggregate this feedback. We present a high-level argument for this claim next and discuss the details later.

\textbf{Three observations about typical RLHF pipelines.} First, granular feedback is sought from individuals. Second, this granular feedback is converted to binary or tertiary feedback before it is aggregated. Third, this coarser form of feedback is not aggregated explicitly. Instead, one trains a neural network using this feedback, and the training process is as if this coarser form of feedback is averaged. More details about these observations are given in Section~\ref{sec:typical_reward_learning}.

\begin{figure}[htp]
    \centering
    \begin{subfigure}{0.45\textwidth}
        \centering
        \includegraphics[width=\textwidth]{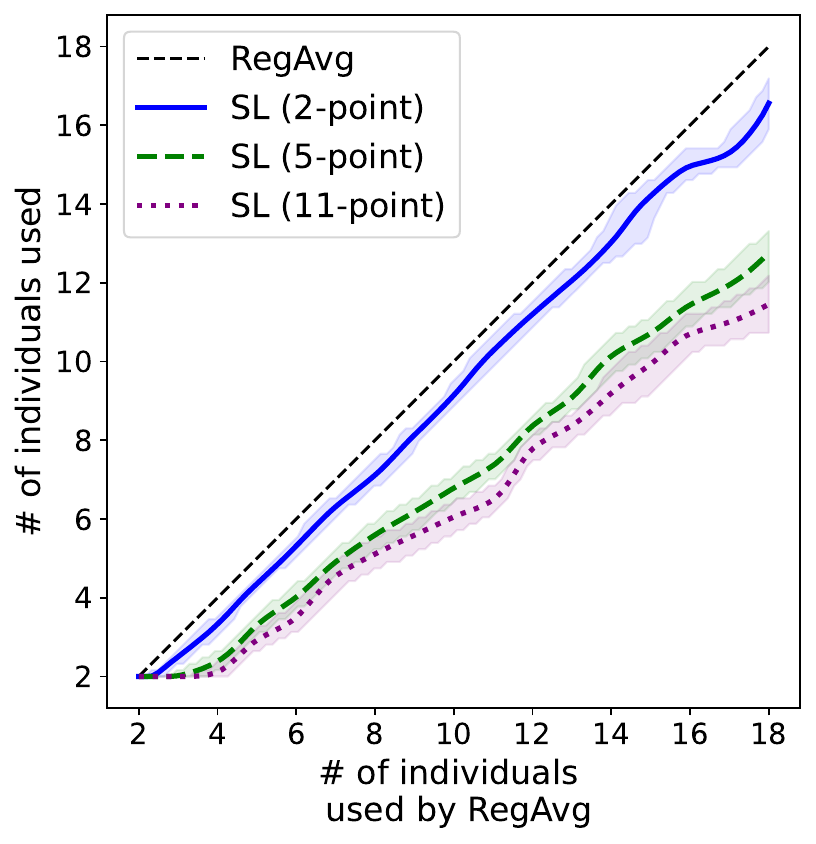}
        \caption{Performance comparison with loss function in Equation~\ref{eq:bin_loss_ignore_neutral}.}
        \label{fig:bin_loss_1}
    \end{subfigure}
    \hfill
    \begin{subfigure}{0.45\textwidth}
        \centering
        \includegraphics[width=\textwidth]{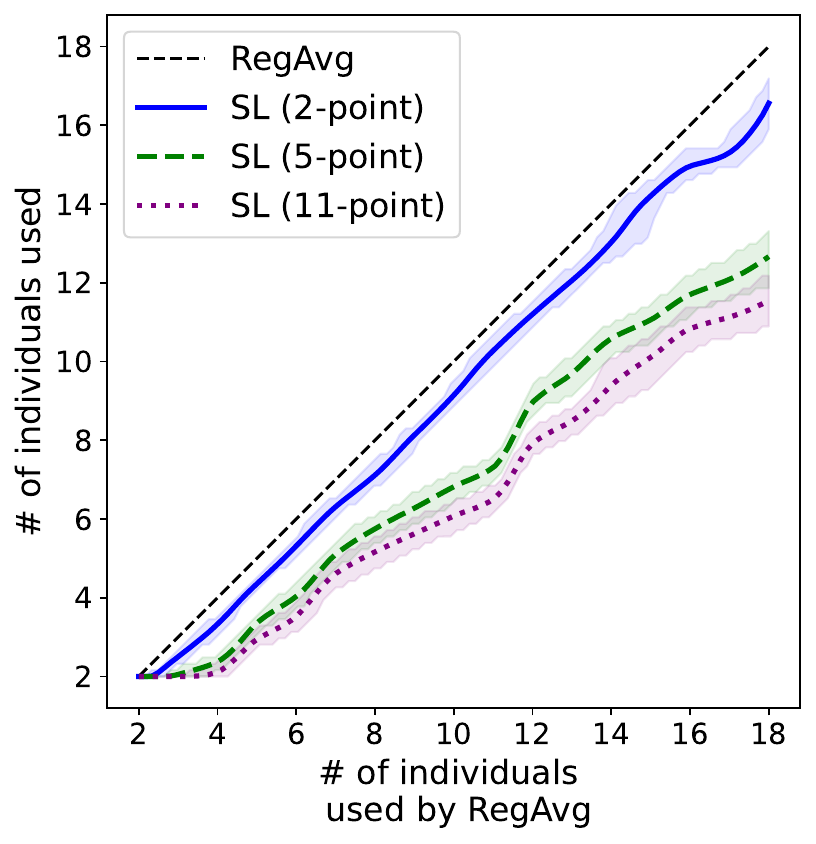}
        \caption{Performance comparison with loss function in Equation~\ref{eq:bin_loss_keep_neutral}.} 
        \label{fig:bin_loss_2}
    \end{subfigure} 
    \caption{Performance comparison for predicting choices of output-set individuals using loss functions in Equations~\ref{eq:bin_loss_ignore_neutral} and \ref{eq:bin_loss_keep_neutral}. For both loss functions, relative to $\regavg$, $\sl$ requires about 36\% and 29\% less individuals with 11-point and 5-point feedback, respectively.}
    \label{fig:perf_comp_bin_losses}
\end{figure}

\textbf{Utility of sophisticated aggregation methods.} We then argue for the following claim. Consider the aggregated feedback in typical RLHF pipelines. If feedback granularity is high, we can match the quality of this aggregated feedback at a reduced cost using sophisticated aggregation methods. To provide empirical evidence for this claim, we consider the experimental setup in Section~\ref{sec:all_evidence}. The only change is in the loss function we consider. Instead of the loss function in Equation~\ref{eq:loss_fn}, we consider loss functions that are more relevant to RLHF (details in Section~\ref{sec:exp_details_bin_loss}). With our experimental setup and these loss functions, we observe that sophisticated aggregation methods like $\sl$ require about 29\% fewer input-set individuals than methods like $\regavg$ when feedback granularity is five; see Figure~\ref{fig:perf_comp_bin_losses}. We discuss later how aggregated feedback in typical RLHF pipelines is related to the output of $\regavg$, but this relation shouldn't be too surprising given the second and third observations above.

Methods like $\sl$ require feedback from output-set individuals to outperform $\regavg$, and collecting this feedback will increase the cost beyond what is implied by results in Figure~\ref{fig:perf_comp_bin_losses}. Despite the extra cost, sophisticated methods can still provide a non-trivial benefit. Here is an example where, despite collecting feedback from output-set individuals, the overall number of scores needed when using $\sl$ is about 20\% less than what would be required in typical pipelines. Suppose we are collecting feedback on task units related to social values and attitudes, and there are 10,000 task units for which we need to aggregate feedback. These task units may be polarizing, so for the $\regavg$ baseline, it might be beneficial to obtain a higher number of scores for each task unit compared to what is typically sought for non-polarizing tasks, such as evaluating article summaries. Assume we get twenty scores for each task unit, and the scores are on a 5-point scale. Suppose, like in Figure~\ref{fig:perf_comp_bin_losses}, after training on 1000 task units, $\sl$ requires 29\% fewer input-set individuals than $\regavg$. Hence, when $\regavg$ uses 20 individuals, $\sl$ requires 14 individuals after training to match the performance of $\regavg$. To train $\sl$, we require scores from 20 additional individuals on the 1000 task units we decided to use for training. We can compute that the total number of scores used by $\regavg$ is 200,000. In contrast, $\sl$ uses a total of 160,000 scores (140,000 from input-set individuals and 20,000 from output-set individuals), thereby requiring 20\% fewer scores than $\regavg$ to produce similar quality of aggregated feedback.

The above example assumes all input-set individuals provide feedback on all task units, but this condition can be relaxed. Since providing feedback for 10,000 task units can be tedious and time-consuming, one must hire more individuals and make different individuals provide feedback on different subsets of task units in parallel. Such feedback collection will lead to missing values in the input because not all individuals will provide feedback for all task units. Depending on missing value frequency, the neural network underlying $\sl$ would require modifications of varying complexity, and training these networks may need larger datasets. For sparse missing values, the following should suffice: appending a binary vector that indicates which individuals' feedback is present. When frequency of missing values is large, one might need to input task unit information and information about the individuals providing feedback for that particular task unit \citep{liu2019learning, gordon2022jury}.

\textbf{Post-aggregation steps.} Finally, to show the main claim of this section, we make the assumption that when we train neural networks using aggregated feedback of similar quality, the trained neural networks generalize similarly to one another. We make this assumption for simplicity. Moreover, a similar assumption is often implicitly made for label aggregation \citep{zhang2016learning, zheng2017truth, zhang2022knowledge}, where aggregation is studied separately from training machine learning models. Now, we discuss how our main claim follows from this assumption and our previous claim on the utility of sophisticated aggregation methods. To do so, we first note that in RLHF, the neural networks directly trained using aggregated feedback are called reward models; more details in Section~\ref{sec:typical_reward_learning}. Our assumption implies that if aggregation methods produce outputs of similar quality, then reward models trained using their outputs will generalize similarly. Moreover, the output of a reward model is one of the two parts of an objective function used in an RLHF pipeline. The other part does not depend on the reward model. LLM parameters are modified to maximize this objective function. If the reward models generalize similarly, the objective functions could guide LLMs to produce similar responses for the same prompts. Hence, it seems plausible to us that the LLMs would have similar quality.

\subsection{Typical procedure to collect and aggregate human feedback}
\label{sec:typical_reward_learning}

We describe how human feedback is collected and aggregated in many major open-source works on RLHF, such as those by \citet{stiennon2020learning, nakano2021webgpt, bai2022training, touvron2023llama, grattafiori2024llama}. The description we provide will make the aforementioned observations more precise.

\textbf{Feedback collection.} The main aspect to remember about the feedback collection step is that the granular feedback is sought from individuals. This step begins with researchers collecting or using an existing dataset of prompts $\{{\rm P}_t \}_{t=1}^T$. For each prompt ${\rm P}_t$, they use an LLM to generate two responses $(\mathrm{R}_{t,1}, \mathrm{R}_{t,2})$. Individuals are hired to give feedback on these responses. Specifically, individual $k$ observes a \textit{comparison triplet} $(\mathrm{P}_t, \mathrm{R}_{t,1}, \mathrm{R}_{t,2})$ and gives feedback $Y_{t,k} \in \mathcal{Y}$. We describe the scoring scales used in the works above and point to specific sections in these works where these details are given:
\begin{itemize}
    \item $\{-L,\cdots,-1, 1,\cdots,L \}$ with $L=4$ is used in the work by \citet{bai2022training} (see Figure~6);
    \item $\{-L,\cdots,-1, 0, 1,\cdots,L \}$ with $L > 1$ is used in \citet{stiennon2020learning} (see Step 2, Appendix C.1), \citet{nakano2021webgpt} (see below Figure 9 in Appendix C.2), \citet{touvron2023llama} (see Section 3.2.1), and \citet{grattafiori2024llama} (see Section 4.2.1).
\end{itemize}
Negative scores indicate preference for ${\rm R}_{t,2}$, positive scores indicate preference for ${\rm R}_{t,1}$, and a zero score indicates no preference. As mentioned earlier, the collected feedback is granular and not binary.

\textbf{Aggregation.} Here, we will make the following two observations more precise: (1) granular feedback is converted to binary or tertiary feedback before it is aggregated, and (2) the aggregation process is as if this coarser feedback is averaged. The aggregation step in RLHF involves training a neural network that takes a prompt-response pair as input and outputs a score. In RLHF terminology, this neural network is referred to as a reward model. We denote it by $r_\theta$ where $\theta$ are the trainable parameters. The aforementioned two observations will become more precise once we state the loss functions used to train the reward models. Before stating the loss functions, we state some notation. We let $\mathcal{W}_t$ denote the individuals who gave feedback on the $t$th comparison triplet, and $\ell: [0,1] \times [0,1] \to [0,\infty)$ denote binary log-loss. Here are the loss functions used to train reward models, along with the works that use these loss functions.
\begin{itemize}
    \item We state a family of loss functions indexed by $y \geq 0$. Let $B_{t,k}^{(y, +)}=\ind[Y_{t,k}>y]$ and $B_{t,k}^{(y, -)}=\ind[Y_{t,k}<-y]$. Then, for each $y$, the loss function is:
    \begin{align*}
        &\mathcal{L}^{(r, 1)}\left( \{ Y_{t,k} \}_{k \in \mathcal{W}_t }, r_\theta({\rm P}_{t}, {\rm R}_{t,1}), r_\theta({\rm P}_{t}, {\rm R}_{t,2})\right) = \ind\left[ \sum_{k \in \mathcal{W}_t} B_{t,k}^{(y, +)} + \sum_{k \in \mathcal{W}_t} B_{t,k}^{(y, -)} > 0 \right]  \\
        &\qquad\qquad\qquad\qquad \times \ell \left( \frac{\sum_{k \in \mathcal{W}_t}B_{t,k}^{(y, +)}}{\sum_{k \in \mathcal{W}_t} B_{t,k}^{(y, +)} + \sum_{k \in \mathcal{W}_t} B_{t,k}^{(y, -)}}, \sigma \left(r_\theta({\rm P}_{t}, {\rm R}_{t,1}) - r_\theta({\rm P}_{t}, {\rm R}_{t,2}) \right) \right),
    \end{align*}
    where $\sigma$ is the sigmoid function. Loosely speaking, the indicator variable above makes the reward model predictions irrelevant where the responses are of similar quality. Loss functions of this type were used for training reward models in the works by \citet{stiennon2020learning} (see Section~3.4), \citet{bai2022training} (see Section~2.2), \citet{touvron2023llama} (see text around Equation~1), and \citet{grattafiori2024llama} (see Section 4.2.1).
    \item Next, consider the following loss function:
    \begin{align*}
        &\mathcal{L}^{(r, 2)}\left(  \{ Y_{t,k} \}_{k \in \mathcal{W}_t }, r_\theta({\rm P}_{t}, {\rm R}_{t,1}), r_\theta({\rm P}_{t}, {\rm R}_{t,2})\right) = \\
        &\ell\left( \frac{\sum_{k \in \mathcal{W}_t} \left( \ind[Y_{t,k} > 0] + 0.5 \ind [Y_{t,k}=0] \right) }{|\mathcal{W}_t|}, \sigma \left(r_\theta({\rm P}_{t}, {\rm R}_{t,1}) - r_\theta({\rm P}_{t}, {\rm R}_{t,2}) \right)  \right).
    \end{align*}
    This loss function is used for training reward models in the work by \citet{nakano2021webgpt} (see Bullet 2, Section~3.2 and Appendix~C.2).
\end{itemize}
We can see from these loss functions that granular feedback is converted to binary or tertiary forms and averaged to train reward models.

For ease of notation, we will refer to quantities that the reward models are being trained to predict as \textit{preference signals}. For the loss functions above, preference signals are $\frac{\sum_{k \in \mathcal{W}_t}B_{t,k}^{(y, +)}}{\sum_{k \in \mathcal{W}_t} B_{t,k}^{(y, +)} + \sum_{k \in \mathcal{W}_t} B_{t,k}^{(y, -)}}$ and $\frac{\sum_{k \in \mathcal{W}_t} \left( \ind[Y_{t,k} > 0] + 0.5 \ind [Y_{t,k}=0] \right) }{|\mathcal{W}_t|}$, respectively.

\subsection{Experimental details}
\label{sec:exp_details_bin_loss}

We provide experimental details relevant to the argument in support of this section's main claim. Specifically, we first state and discuss loss functions relevant to Figure~\ref{fig:perf_comp_bin_losses}. Then, we highlight the key difference in datasets commonly used for RLHF and the dataset we used in our experiments. We also provide a connection between preference signals used in typical RLHF pipelines and the output of $\regavg$. Later, we discuss how exactly the plots in Figure~\ref{fig:perf_comp_bin_losses} are constructed, and why the results could generalize beyond the current experimental setup.

\textbf{Loss functions.} We state and discuss two loss functions that will be used to measure the quality of preference signals produced by different methods. Suppose for each comparison triplet $({\rm P}_t, {\rm R}_{t,1},{\rm R}_{t,2})$, we collected feedback from a large population of individuals. Loosely speaking, we are interested in calculating the fraction of the population that prefers response ${\rm R}_{t,1}$ over response ${\rm R}_{t,2}$. To formalize this notion, we first recall some notation introduced in previous sections. We denote the population CDF by $P_t$ (see Equation~\ref{eq:pop_cdf}) and a prediction for $P_t$ as $Q_t$. Let the corresponding PMFs be $p_t$ and $q_t$, respectively. Furthermore, for simplicity of notation, if $0$ is not in the scoring scale $\mathcal{Y}$, we let $p_t(0)=0$ and $q_t(0)=0$. Finally, here are the loss functions we consider:
\begin{subequations}
\begin{align}
    \label{eq:bin_loss_ignore_neutral}
    \begin{aligned}
    \mathcal{L}^{(p, 1)}(p_t, q_t) = \ind\left[\sum_{y \in \mathcal{Y}; y \neq 0} p_t(y) > 0 \right] \ell \left(\frac{\sum_{y \in \mathcal{Y}; y>0} p_t(y)}{\sum_{y \in \mathcal{Y}; y \neq 0} p_t(y)},  \frac{\sum_{y \in \mathcal{Y}; y>0} q_t(y)}{\sum_{y \in \mathcal{Y}; y \neq 0} q_t(y)} \right) \\
    + \ind\left[\sum_{y \in \mathcal{Y}; y \neq 0} p_t(y) = 0 \right] \left(-p_t(0) \log q_t(0) \right);
    \end{aligned} \\
    \label{eq:bin_loss_keep_neutral}
    \mathcal{L}^{(p, 2)}(p_t, q_t) = \ell \left(\sum_{y \in \mathcal{Y}; y>0} p_t(y) + 0.5 p_t(0), \sum_{y \in \mathcal{Y}; y>0} q_t(y) + 0.5 q_t(0) \right).
\end{align}
\end{subequations}
We now provide some intuition for these loss functions. First, note that $\sum_{y \in \mathcal{Y}; y>0} p_t(y)$ is equal to the fraction of individuals in the population that prefer ${\rm R}_{t,1}$ over ${\rm R}_{t,2}$. This sum appears in both loss functions. Also, if there is no neutral feedback, i.e., if $p_t(0)=q_t(0)=0$, these loss functions take the same values, and the sum above dictates what the loss values are. Moreover, these loss functions are inspired by loss functions $\mathcal{L}^{(r,1)}$ and $\mathcal{L}^{(r,2)}$ that were used for training reward models.

\textbf{Dataset.} Our experiments are not on a dataset with comparison triplets but on a dataset related to social surveys. See Section~\ref{sec:dataset} for more details about our dataset and an explanation for why we did not use existing RLHF datasets.

\textbf{Why $\regavg$ as baseline?} It is a simple exercise to check that preference signals given in Section~\ref{sec:typical_reward_learning} can be derived using the empirical CDF of the feedback given by the hired individuals. Furthermore, by definition, the empirical CDF is the same as $\regavg$'s predictive CDF when the regularization parameter $\gamma$ is zero. This equivalence suggests that we should use $\regavg$ with $\gamma=0$ as our baseline. However, this would be a strawman. While the preference signals in Section~\ref{sec:typical_reward_learning} have not been regularized, regularization techniques such as weight decay and early stopping are almost always employed to train reward models. While we don't train reward models, we aim to approximate the effect of regularization in the training process, which we achieve by tuning the regularization hyperparameter $\gamma$. To tune $\gamma$, we use the same tuning procedure as given in Section~\ref{sec:policies}.

\textbf{Performance comparison.} For each of the two loss functions, we do the following. We first vary the number of individuals $K$. Then, for each $K$, we compute the average loss suffered by $\regavg$ and $\sl$ when they use $K$ individuals as input. We find the raw loss values hard to interpret, so we have moved them to Appendix~\ref{app:bin_loss_values}. Instead, we plot the number of individuals $\sl$ requires to match the performance of $\regavg$ for different performance levels. These plots are given in Figure~\ref{fig:perf_comp_bin_losses}. We can see that there is a clear benefit in aggregating granular feedback using $\sl$. Specifically, $\sl$ requires about 36\% and 29\% fewer individuals compared to $\regavg$ with 11-point and 5-point feedback, respectively. While the performance of $\sl$ is slightly better with 11-point feedback compared to the performance with 5-point feedback, it might still be better to collect 5-point feedback in other scenarios, given that it is much simpler and is ubiquitous across applications.

\textbf{Will the results in Figure~\ref{fig:perf_comp_bin_losses} generalize?} It is hard to predict the magnitude of the advantage of $\sl$ relative to $\regavg$ for these loss functions with any precision. However, some qualitative properties of the advantage of $\sl$ seem clear. Specifically, the advantage of $\sl$ should increase with feedback granularity, and even for feedback granularity as large as 5, the advantage may be large. In Section~\ref{sec:intuition}, we argued for a similar claim for the loss function in Equation~\ref{eq:loss_fn}. This argument can be easily adapted to support a similar claim for the loss functions presented in this section. We skip a detailed argument, but provide an overview below. $\regavg$ converts the granular feedback into binary or tertiary feedback before using it to produce preference signals. This is because the preference signals produced by $\regavg$ are regularized versions of preference signals in Section~\ref{sec:typical_reward_learning}. Converting granular feedback into coarser feedback loses information. This loss is likely to be more significant for feedback of greater granularity. Under certain conditions, such as having a sufficiently large dataset and a sufficiently large neural network, methods like $\sl$ can extract all the information from their inputs that is useful in making predictions. Hence, with more granular feedback, $\sl$ has more information that it can exploit to outperform $\regavg$. Therefore, the claim follows.

\section{Limitations and avenues for future work}
\label{sec:limitations}

\textbf{Advantage not shown for unsupervised methods.} Our work draws on ideas from the field of RLHF but is primarily related to the field of aggregation algorithms. A reader who has worked in this field would point out that aggregation algorithms are typically unsupervised \citep{zhang2016learning, zheng2017truth, zhang2022knowledge}, and we have not demonstrated our main claim for such algorithms. We leave it for future work to do such a demonstration. Nonetheless, our results are informative. At least for our task units, it is clear that sophisticated unsupervised algorithms will not offer any advantage for binary data, but could potentially offer a large advantage when the feedback is more granular.

\textbf{Experimental evaluation on one dataset only.} We demonstrated that the advantage of $\sl$ increases with granularity only on one dataset. While the argument in Section~\ref{sec:intuition}, provides reason to believe our results generalize to other datasets, future works should perform more experimental evaluations. However, we would like to remind that, as argued in Section~\ref{sec:reason_new_dataset}, existing datasets may be unsuitable for performing the kind of analysis we performed in this work. Hence, to perform more experimental evaluations, one might have to collect new datasets.

\textbf{Advantage unclear when there are missing values.} By missing values, we mean that not all individuals give feedback on all task units. While there were no missing values in our experiments, in practice, this is often the case. We very briefly discussed how missing values could be handled in the introductory paragraphs of Section~\ref{sec:rlhf}. However, more research is needed to understand how the advantage of sophisticated aggregation algorithms changes with the amount of missing values.

\textbf{Advantage may not keep increasing with granularity.} We demonstrated that for our task units, the advantage of $\sl$ increases with granularity for scoring scales that contain between two and eleven scores. However, if we kept increasing granularity, the advantage might not increase. There may be many reasons, but one likely reason is that the workers may not be able to meaningfully use a very granular scale. For example, we might give them a hundred-point scale, but they might only answer in increments of five or ten. We leave it for future work to conduct a systematic analysis on how granular the scales can be before workers stop using the full rating scale. 

\textbf{Broader impact.} We collected a dataset on social attitudes and values in Thailand, and will release it. All the workers who gave feedback are from Thailand. This could allow unscrupulous people to utilize this data to develop systems that discriminate against Thai people. However, we think the more common scenario is that people utilize our data in a way that leads to more positive outcomes for everyone. Therefore, we will release the data.

\section*{Acknowledgements}
This research was supported by a seed fund from the Stanford Institute for Human-Centered AI, contributed by SCBX and SCB 10X. It was also supported by Grant W911NF2410095 from the US Army Research Office. We thank Adisai Na-Thalang and Chanakan Wittayasakpan from SCB 10X for their assistance in collecting crowdworker feedback. The paper also benefited tremendously from discussions with Saurabh Kumar, Jonathan Colaco Carr, and Wanqiao Xu. Finally, we thank all the crowdworkers, as without them, this research would not have been accomplished.

\bibliography{references}

\appendix

\section{Dataset creation details}
\label{app:dataset_creation}
Here, we give details about why and how we created our dataset. First, give more details on why we couldn't use existing datasets. Then, we describe how the task units were created. Finally, we discuss how we got feedback from individuals and the quality checks we performed.

\subsection{Reasons for collecting a new dataset}
\label{app:reason_new_dataset}
Our decision to collect a new dataset was primarily influenced by the kind of analysis we wanted to perform. We were interested in investigating the following. One, the performance of different policies, as there is an increase in the number of individuals giving feedback. Two, the change in their performance as we increase feedback granularity. Based on these objectives and some preliminary analysis, we determined that about forty individuals should provide feedback for each task unit, and the granularity of the feedback $|\mathcal{Y}|$ should be around ten. We also wanted the individuals to give feedback without seeing each other's answers. Given these requirements, we couldn't use existing datasets. Specifically, popular RLHF datasets \citep{awesome_llm_human_preference_datasets} and some datasets \citep{TREC-RF10-Crowd, zhou2012learning} used in crowdsourcing literature have too few individuals giving feedback for each question. Specifically, each task unit has fewer than six individuals giving feedback. Other datasets used in the crowdsourcing literature, like those by \citet{AdultContent3-HCOMP2010} and \citet{halpern2016mobile}, and a recent dataset by \citet{santurkar2023whose} based on opinion surveys do not meet our requirement on granularity because they get feedback on four or five-point scales. Finally, the dataset by \citet{MovieLens32M} consists of user ratings from a movie recommendation website. The ratings are on a ten-point scale, and many movies have over forty ratings. However, users can see previous ratings before generating their own. Hence, we didn't use this dataset.

\subsection{Creating task units}
\label{app:ques_gen}

\noindent \textbf{Topic selection.} We first scraped a list of topics from the website of Pew Research Center \citep{pew_research_topics}. In total, there are 420 topics. Here are a few examples: Demographics \& Politics, Internet \& Technology, and Economic Conditions. Subsequently, we prompted Claude 3.5 Sonnet (\texttt{claude-3-5-sonnet-20241022}) to rate each topic based on a few criteria. The idea was to rate topics so we could form questions using them that are not too personal, require less specialized knowledge, and are concrete. The prompt and system prompt we used to rate these topics are in Figure~\ref{pg:topic-prompt}. After generating the ratings, we retain only topics with a rating of three or more out of five, yielding 301 topics.

\vspace{1.5mm}
\noindent \textbf{Generation.} We used Claude 3.5 Sonnet to generate 10 task units for each of the 301 topics. Each generated task unit consists of (i) a context, (ii) a question, and (iii) a textual description of what -5, 0, and +5 scores indicate. The specific prompt and system prompt we used are given in Figure~\ref{pg:question-prompt} with the generation temperature parameter set to 0.2. This yielded 3010 task units in total.

\vspace{1.5mm}
\noindent \textbf{Filtering and refinement.} We filtered task units by prompting Claude Sonnet 3.5 to assign a score to each task unit on its clarity, relevance to Thai residents, factual accuracy, and concreteness. Subsequently, thousand task units with the highest scores were selected. We observed that the textual description associated with the score of ``0" for some task units was not necessarily neutral. So, we refined them to ensure a neutral description. Moreover, after collecting feedback from crowdworkers, we realized that there are 998 unique units instead of a thousand. While repeated questions can be undesirable in certain settings, we found them useful in conducting quality checks. We give more details in the next subsection.

\begin{figure}[htp]
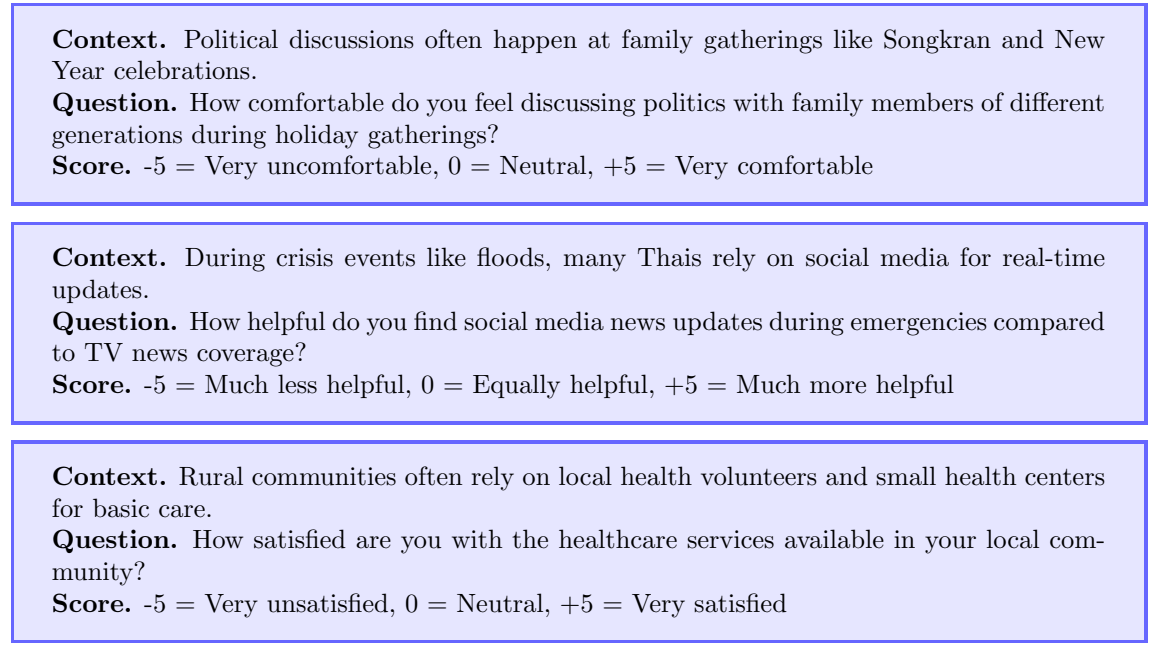

\begin{minipage}{\textwidth}
\centering
\begin{contextbox}
\textbf{Context.}
Political discussions often happen at family gatherings like Songkran and New Year celebrations.

\textbf{Question.} How comfortable do you feel discussing politics with family members of different generations during holiday gatherings?

\textbf{Score.} -5 = Very uncomfortable, 0 = Neutral, +5 = Very comfortable
\end{contextbox}

\begin{contextbox}
\textbf{Context.}
During crisis events like floods, many Thais rely on social media for real-time updates.

\textbf{Question.} How helpful do you find social media news updates during emergencies compared to TV news coverage?

\textbf{Score.} -5 = Much less helpful, 0 = Equally helpful, +5 = Much more helpful
\end{contextbox}

\begin{contextbox}
\textbf{Context.}
Rural communities often rely on local health volunteers and small health centers for basic care.

\textbf{Question.} How satisfied are you with the healthcare services available in your local community?

\textbf{Score.} -5 = Very unsatisfied, 0 = Neutral, +5 = Very satisfied
\end{contextbox}

\end{minipage}
\caption{Examples of a task unit that we gave to the crowdworkers.}
\label{pg:example_task_units2}
\end{figure}

\vspace{1.5mm}
\noindent \textbf{Examples of task units.} Please see Figure~\ref{pg:example_task_units2}.

\newtcolorbox{promptbox}{
    colback=red!10, %
    colframe=red!60, %
    boxrule=0.5mm,  %
    width=\textwidth,
    sharp corners
}

\newtcolorbox{systempromptbox}{
    colback=green!10, %
    colframe=green!60, %
    boxrule=0.5mm,  %
    width=\textwidth,
    sharp corners
}

\begin{figure}[htp]
\begin{minipage}{\textwidth}
\centering
\begin{promptbox}
\begin{lstlisting}
Review this PEW Research topic: {topic} and evaluate it based on the following criteria:
1. Focus on systems/policies rather than personal beliefs
2. Can be discussed using concrete metrics (costs, percentages, numbers)
3. Are relevant across different cultural contexts
4. Don't require specialized knowledge
5. Can be framed without personal disclosure
6. Have observable real-world implications

For this topic:
- Rate 1-5 on how well it meets these criteria
- Note specific metrics/numbers that could anchor questions
- Flag if the topic needs cultural adaptation for Thailand

Give lower ratings to topics that:
- Require religious/ideological positions
- Focus on US-specific issues
- Need extensive background knowledge
- Ask about personal practices/beliefs

To make it easier to process, please provide feedback in JSON format. 
The key-value pairs should be the following.
Topic: {item}
Rating: [1-5]
Metrics: [specific metrics]
Adaptation: [yes/no]

All your feedback must be in a single JSON object. Otherwise, my parsing code will fail.
\end{lstlisting}
\end{promptbox}

\begin{systempromptbox}
\begin{lstlisting}
You are evaluating public policy topics for a large-scale survey in Thailand. 
You focus on:
- Topics that can be discussed using concrete metrics (costs, percentages, numbers)
- System and policy-level issues over individual behaviors
- Issues with observable societal impacts
- Universal concepts that exist across countries
You exclude topics that:
- Require personal disclosure or beliefs
- Need specialized knowledge
- Are primarily US-centric
- Focus on religion/ideology
\end{lstlisting}
\end{systempromptbox}
\end{minipage}
\caption{Prompt and system prompt for rating topics. The topic list is scraped from the website of Pew Research Center \citep{pew_research_topics}. The prompts are written with the idea of choosing topics that lead to questions that are not too personal, are concrete, and require less specialized knowledge.}
\label{pg:topic-prompt}
\end{figure}

\begin{figure}[htp]
\begin{minipage}{\textwidth}
\centering
\begin{promptbox}
\begin{lstlisting}
Generate 10 survey questions about {topic} for Thai workers. Each question must:
1. Use everyday language familiar to Thai workers but address complex aspects of the topic
2. Include specific examples, numbers, or scenarios relevant to Thai society and daily life
3. Focus on concrete comparisons or changes over time in a Thai context
4. Reference observable real-world situations that Thai workers would encounter
5. Avoid academic jargon but don't oversimplify the underlying concepts
6. Consider Thai cultural values and social norms when framing sensitive topics
7. Use examples from urban and rural Thai settings when applicable
8. Use a scale from -5 to +5

If needed, include a brief context statement before the question. Please generate questions in English, but ensure they are easily translatable into Thai. Also, give the questions in JSON format. Do not include any other text as it breaks my parser.
Format each question as follows:
{{"context": "Example context using specific facts or scenarios relevant to Thailand",
"question": "The actual question focusing on a specific aspect or implication",
 "scale": "Rate from -5 to +5."}}.
In the value for the "scale" key, please explain what -5, 0 and +5 mean. For example, "-5 = Strongly disagree, 0 = Neither agree nor disagree, +5 = Strongly agree". Format the questions as a list of dictionaries.
\end{lstlisting}
\end{promptbox}

\begin{systempromptbox}
\begin{lstlisting}
You are an expert survey methodologist who specializes in writing clear questions for the general public.
\end{lstlisting}
\end{systempromptbox}
\end{minipage}
\caption{Prompt and system prompt for generating task units. The idea is to generate task units with questions that are easy to understand but target aspects where people often have conflicting experiences and opinions.}
\label{pg:question-prompt}
\end{figure}

\subsection{Crowdworker feedback}
\label{app:collecting_worker_feedback}
\noindent \textbf{Collecting crowdworker feedback} We recruited 40 crowdworkers from a Thai online freelance community, primarily students and freelancers. Each of them gave feedback for a total of 1020 units. In addition to the thousand task units mentioned earlier, we repeated twenty task units to be able to perform quality checks. We will give more details slightly later. The task units were presented in five Google forms, allowing crowdworkers to save their progress and complete them at their own pace. They were given approximately one week to finish the task. The instructions were simple and are stated in Figure~\ref{pg:instructions}. The task units were all translated into Thai for the crowdworkers. The crowdworkers were compensated for this task at a rate more than three times the minimum wage in Thailand. The crowdworkers were informed that their responses would be used for research purposes. The data collection was reviewed and approved by the management, and we followed internal ethical guidelines.

\begin{figure}[htp]
\begin{minipage}{\textwidth}
\centering
\begin{promptbox}
\begin{itemize}
    \item Please answer the questionnaire based on your true feelings, rating your opinion on a scale from -5 to 5. If you feel neutral, select 0.
    \item If you do not have an experience that matches the situation asked, please try to assume or compare as close as possible so that you can score.
    \begin{itemize}
        \item For example, if the question is “How has your child’s ability to pay for school fees changed over the past year?”, you can assume that you have children and predict how your income will change. Or, if you are still studying, you can compare how your ability to pay for school fees has changed and score accordingly.
    \end{itemize}
    \item The number of questions that you choose 0 for each section should not exceed 20 questions or a total of no more than 100 questions.
\end{itemize}
\end{promptbox}
\end{minipage}
\caption{Instructions for the crowdworkers.}
\label{pg:instructions}
\end{figure}

\vspace{1.5mm}
\noindent \textbf{Quality checks.} First, as seen in Figure~\ref{pg:instructions}, we instructed these crowdworkers not to answer more than ten percent of the task units with a zero score. Most crowdworkers followed this: only one of the forty crowdworkers we hired did not follow this instruction. Hence, we did not accept their data. Second, we repeated some of the task units. Specifically, twenty-two task units are repeated out of the thousand unique task units we asked. The motivation for repeating some of the task units was understanding how `noisy' the crowdworkers were. If a crowdworker's feedback changes a lot for the same task unit, then it is intuitive to call this crowdworker more noisy. While we did find some crowdworkers noisy, we did not exclude any data. Now, we will present some key statistics about changes in crowdworkers' feedback. Let $\mathcal{I}=\{(\tau, \tau'): \tau < \tau' \text{ and } Q_\tau=Q_{\tau'}\}$. Loosely speaking, $\mathcal{I}$ contains pairs of indices of repeated task units. Using the repeated units, we compute the mean feedback change $\Delta_k = \sum_{(\tau, \tau') \in \mathcal{I}} |Y_{\tau,k}-Y_{\tau',k}|/|\mathcal{I}|$ for each crowdworker $k$. Please see Table~\ref{tab:mean_pref_change_stats} for some statistics of $\Delta_k$. An example statistic: the mean, $\nicefrac{\sum_{k=1}^{K''} \Delta_k}{K''}$, where $K''$ is the number of crowdworkers in our dataset and is equal to thirty-nine.

\begin{table}[htp]
    \centering
    \begin{tabular}{|c|c|}
    \hline
        \begin{tabular}{c}
        \\[-0.2cm]
         \textbf{statistic} \\[0.2cm]
        \end{tabular}
        &
        \begin{tabular}{c}
        \\[-0.2cm]
        \textbf{value} \\[0.2cm]
        \end{tabular} \\
        \hline
        
        \begin{tabular}{c}
        \\[-0.2cm]
        mean
        \\[0.2cm]
        \end{tabular}
        &
        \begin{tabular}{c}
        \\[-0.2cm]
        1.54
        \\[0.2cm]
        \end{tabular} \\
        \hline
        
        \begin{tabular}{c}
        \\[-0.2cm]
        standard deviation
        \\[0.2cm]
        \end{tabular}
        &
        \begin{tabular}{c}
        \\[-0.2cm]
        0.74
        \\[0.2cm]
        \end{tabular} \\
        \hline
        
        \begin{tabular}{c}
        \\[-0.2cm]
        minimum 
        \\[0.2cm]
        \end{tabular}
        &
        \begin{tabular}{c}
        \\[-0.2cm]
        0.45
        \\[0.2cm]
        \end{tabular} \\
        \hline

        \begin{tabular}{c}
        \\[-0.2cm]
        25$^{\text{th}}$ percentile
        \\[0.2cm]
        \end{tabular}
        &
        \begin{tabular}{c}
        \\[-0.2cm]
        1.04
        \\[0.2cm]
        \end{tabular} \\
        \hline

        \begin{tabular}{c}
        \\[-0.2cm]
        median
        \\[0.2cm]
        \end{tabular}
        &
        \begin{tabular}{c}
        \\[-0.2cm]
        1.45
        \\[0.2cm]
        \end{tabular} \\
        \hline
        
        \begin{tabular}{c}
        \\[-0.2cm]
        75$^{\text{th}}$ percentile
        \\[0.2cm]
        \end{tabular}
        &
        \begin{tabular}{c}
        \\[-0.2cm]
        1.84
        \\[0.2cm]
        \end{tabular} \\
        \hline

        \begin{tabular}{c}
        \\[-0.2cm]
        maximum
        \\[0.2cm] 
        \end{tabular}
        &
        \begin{tabular}{c}
        \\[-0.2cm]
        4.18
        \\[0.2cm]
        \end{tabular} \\
        \hline
    \end{tabular} \\[4pt]
    \caption{Statistics of mean feedback change $\Delta_k$. We had asked crowdworkers to report feedback in $\{-5,-4,\cdots,4,5\}$. The dataset is moderately noisy. Half the crowdworkers change their feedback between 1 and 2 points, i.e., 10 and 20\% of the maximum possible change of 10 points. Moreover, there is non-trivial variation in noise levels: loosely speaking, one crowdworker's feedback changes by less than half a point while another crowdworker's feedback changes by more than four points.}
    \label{tab:mean_pref_change_stats}
\end{table}

\section{Data preparation details}
\label{app:data_prep_details}

\textbf{Dividing the crowdworkers into input-set and output-set.} Recall that we have thirty-nine crowdworkers giving feedback on each task unit of our dataset. Also, recall that we split the crowdworkers into two sets, the input-set and the output-set. The sizes of these sets are twenty and nineteen, respectively. The details about their roles are given in Section~\ref{sec:data_prep}. Here, we explain how we divided the crowdworkers into these sets. We used the NumPy random number generator with seed forty-two to sample nineteen indices without replacement. The crowdworkers at these indices were put in the output-set, and the rest were put in the input-set. See \verb|data\create_dataset_11_pt.py| in the code repository.

\textbf{Creating the 5-point dataset.} Recall that our original dataset gets feedback from crowdworkers on an 11-point scale. We want to maintain proportionality. What this means here is that each score on the 5-point scale should be mapped to 2.2 (=11/5) scores on the 11-point scale. As 2.2 is not an integer, some scores on the 11-point scale need to be divided probabilistically. Here is what we mean. Consider Table~\ref{tab:prob_11_to_5}. It contains probabilities for how feedback on the 11-point scale will get mapped to 5-point scale. For example, if a crowdworker's feedback is -3 on the 11-point scale, then with probabilities 0.2 and 0.8, it will get mapped to -2 and -1, respectively. You can check that each score on the 5-point scale is mapped to 2.2 scores on the 11-point scale. The details about the random number generator are available in \verb|data\create_dataset_5_pt.py| in the code repository.

\begin{table}[htp]
    \centering
    \begin{tabular}{|c|c|c|c|c|c|}
    \toprule
    & \multicolumn{5}{c|}{\textbf{5-point Scale}} \\
    \cmidrule{2-6}
    \textbf{11-point Scale} & \textbf{-2} & \textbf{-1} & \textbf{0} & \textbf{1} & \textbf{2} \\
    \midrule
    \textbf{-5} & 1.0 & & & & \\
    \textbf{-4} & 1.0 & & & & \\
    \textbf{-3} & 0.2 & 0.8 & & & \\
    \textbf{-2} & & 1.0 & & & \\
    \textbf{-1} & & 0.4 & 0.6 & & \\
    \textbf{0} & & & 1.0 & & \\
    \textbf{1} & & & 0.6 & 0.4 & \\
    \textbf{2} & & & & 1.0 & \\
    \textbf{3} & & & & 0.8 & 0.2 \\
    \textbf{4} & & & & & 1.0 \\
    \textbf{5} & & & & & 1.0 \\
    \bottomrule
    \end{tabular}
    \caption{Probability mapping from scores on the 11-point scale to the scores on the 5-point scale. Here is an example to clarify how this mapping works. If a crowdworker's feedback is -3 on the 11-point scale, then with probabilities 0.2 and 0.8, it will get mapped to -2 and -1, respectively.}
    \label{tab:prob_11_to_5}
\end{table}

\section{Details of modified MLP, training process, and hyperparameter tuning}
\label{app:sl_hyperparam}

We use a modified MLP because we are predicting a CDF. The elements of a CDF are monotonically increasing, but the elements of a standard MLP's output may not increase monotonically. Hence, we modified it by applying softmax to the output of the last layer to get a valid PMF and then transforming it to a CDF. We train this MLP using standard deep learning regularization and optimization techniques: AdamW optimizer with dropout, batch normalization, weight decay, and early stopping.

We use the same set of hyperparameters for all values of $K$ to reduce the hyperparameter tuning we need. The quantities that we tuned are the following: the number of hidden layers, the size of the hidden layers, the batch size, the learning rate, the weight decay, and the patience parameter for early stopping. We could have also tuned the dropout probability, but we just fixed it to $0.2$. Moreover, we did not do a grid search but tried to improve performance step by step. For example, if increasing the hidden layer's size improved the performance, we would increase it further until there was no observable difference. Tables \ref{tab:hyperparam_2}, \ref{tab:hyperparam_5}, and \ref{tab:hyperparam_11} have the hyperparameter values we tried and the values we chose for the cases with 2-point, 5-point, and 11-point feedback, respectively.

\textbf{Compute resources used.} We had access to one n2-highcpu-64 machine on Google Cloud. This machine has 64 cores and a total RAM of 64GB. We equipped it with a 100GB disk. Our experiments can be parallelized. After hyperparameter tuning, they take less than 3 minutes to complete on the previously mentioned machine. We do not have an exact number for the amount of resources we spent on hyperparameter tuning but it should definitely take less than an hour. 

We had performed a simulation study before collecting the dataset that we have presented here. This would have used a few hours of compute time on the aforementioned machine. We are choosing to not release the details of the simulation study as we are not sure how much value it adds to the current paper.

\begin{table}[htp]
    \centering
    \begin{tabular}{|c|c|c|}
    \hline
        \begin{tabular}{c}
        \\[-0.2cm]
         \textbf{hyperparameter} \\[0.2cm]
        \end{tabular}
        &
        \begin{tabular}{c}
        \\[-0.2cm]
        \textbf{all values} \\[0.2cm]
        \end{tabular}
        &
        \begin{tabular}{c}
        \\[-0.2cm]
        \textbf{chosen value} \\[0.2cm]
        \end{tabular} \\
        \hline
        
        \begin{tabular}{c}
        \\[-0.2cm]
        number of hidden layers
        \\[0.2cm]
        \end{tabular}
        &
        \begin{tabular}{c}
        \\[-0.2cm]
        $\{1, 2\}$
        \\[0.2cm]
        \end{tabular}
        &
        \begin{tabular}{c}
        \\[-0.2cm]
        $2$
        \\[0.2cm]
        \end{tabular} \\
        \hline
        
        \begin{tabular}{c}
        \\[-0.2cm]
        size of both hidden layers
        \\[0.2cm]
        \end{tabular}
        &
        \begin{tabular}{c}
        \\[-0.2cm]
        $\{15, 25, 50\}$
        \\[0.2cm]
        \end{tabular}
        &
        \begin{tabular}{c}
        \\[-0.2cm]
        $25$
        \\[0.2cm]
        \end{tabular} \\
        \hline
        
        \begin{tabular}{c}
        \\[-0.2cm]
        batch size
        \\[0.2cm]
        \end{tabular}
        &
        \begin{tabular}{c}
        \\[-0.2cm]
        $\{25, 50\}$
        \\[0.2cm]
        \end{tabular}
        &
        \begin{tabular}{c}
        \\[-0.2cm]
        $25$
        \\[0.2cm]
        \end{tabular} \\
        \hline
        
        \begin{tabular}{c}
        \\[-0.2cm]
        learning rate
        \\[0.2cm]
        \end{tabular}
        &
        \begin{tabular}{c}
        \\[-0.2cm]
        $\{0.005, 0.01 \}$
        \\[0.2cm]
        \end{tabular}
        &
        \begin{tabular}{c}
        \\[-0.2cm]
        $0.01$
        \\[0.2cm]
        \end{tabular} \\
        \hline
        
        \begin{tabular}{c}
        \\[-0.2cm]
        weight decay
        \\[0.2cm]
        \end{tabular}
        &
        \begin{tabular}{c}
        \\[-0.2cm]
        $\{0.0005, 0.001 \}$
        \\[0.2cm]
        \end{tabular}
        &
        \begin{tabular}{c}
        \\[-0.2cm]
        $0.001$
        \\[0.2cm]
        \end{tabular} \\
        \hline
        
        \begin{tabular}{c}
        \\[-0.2cm]
        patience
        \\[0.2cm]
        \end{tabular}
        &
        \begin{tabular}{c}
        \\[-0.2cm]
        $\{4, 8\}$
        \\[0.2cm]
        \end{tabular}
        &
        \begin{tabular}{c}
        \\[-0.2cm]
        $8$
        \\[0.2cm]
        \end{tabular} \\
        \hline
        
    \end{tabular} \\[4pt]
    \caption{Hyperparameter values for the case with 2-point feedback.}
    \label{tab:hyperparam_2}
\end{table}

\begin{table}[htp]
    \centering
    \begin{tabular}{|c|c|c|}
    \hline
        \begin{tabular}{c}
        \\[-0.2cm]
         \textbf{hyperparameter} \\[0.2cm]
        \end{tabular}
        &
        \begin{tabular}{c}
        \\[-0.2cm]
        \textbf{all values} \\[0.2cm]
        \end{tabular}
        &
        \begin{tabular}{c}
        \\[-0.2cm]
        \textbf{chosen value} \\[0.2cm]
        \end{tabular} \\
        \hline
        
        \begin{tabular}{c}
        \\[-0.2cm]
        number of hidden layers
        \\[0.2cm]
        \end{tabular}
        &
        \begin{tabular}{c}
        \\[-0.2cm]
        $\{1,2 \}$
        \\[0.2cm]
        \end{tabular}
        &
        \begin{tabular}{c}
        \\[-0.2cm]
        $2$
        \\[0.2cm]
        \end{tabular} \\
        \hline
        
        \begin{tabular}{c}
        \\[-0.2cm]
        size of both hidden layers
        \\[0.2cm]
        \end{tabular}
        &
        \begin{tabular}{c}
        \\[-0.2cm]
        $\{15, 25, 50 \}$
        \\[0.2cm]
        \end{tabular}
        &
        \begin{tabular}{c}
        \\[-0.2cm]
        $25$
        \\[0.2cm]
        \end{tabular} \\
        \hline
        
        \begin{tabular}{c}
        \\[-0.2cm]
        batch size
        \\[0.2cm]
        \end{tabular}
        &
        \begin{tabular}{c}
        \\[-0.2cm]
        $\{25, 50, 100 \}$
        \\[0.2cm]
        \end{tabular}
        &
        \begin{tabular}{c}
        \\[-0.2cm]
        $50$
        \\[0.2cm]
        \end{tabular} \\
        \hline
        
        \begin{tabular}{c}
        \\[-0.2cm]
        learning rate
        \\[0.2cm]
        \end{tabular}
        &
        \begin{tabular}{c}
        \\[-0.2cm]
        $\{0.001, 0.005, 0.01 \}$
        \\[0.2cm]
        \end{tabular}
        &
        \begin{tabular}{c}
        \\[-0.2cm]
        $0.01$
        \\[0.2cm]
        \end{tabular} \\
        \hline
        
        \begin{tabular}{c}
        \\[-0.2cm]
        weight decay
        \\[0.2cm]
        \end{tabular}
        &
        \begin{tabular}{c}
        \\[-0.2cm]
        $\{0.0001, 0.0005, 0.001 \}$
        \\[0.2cm]
        \end{tabular}
        &
        \begin{tabular}{c}
        \\[-0.2cm]
        $0.001$
        \\[0.2cm]
        \end{tabular} \\
        \hline
        
        \begin{tabular}{c}
        \\[-0.2cm]
        patience
        \\[0.2cm]
        \end{tabular}
        &
        \begin{tabular}{c}
        \\[-0.2cm]
        $\{4, 8\}$
        \\[0.2cm]
        \end{tabular}
        &
        \begin{tabular}{c}
        \\[-0.2cm]
        $8$
        \\[0.2cm]
        \end{tabular} \\
        \hline
        
    \end{tabular}\\[4pt]
    \caption{Hyperparameter values for the case with 5-point feeedback.}
    \label{tab:hyperparam_5}
\end{table}

\begin{table}[htp]
    \centering
    \begin{tabular}{|c|c|c|}
    \hline
        \begin{tabular}{c}
        \\[-0.2cm]
         \textbf{hyperparameter} \\[0.2cm]
        \end{tabular}
        &
        \begin{tabular}{c}
        \\[-0.2cm]
        \textbf{all values} \\[0.2cm]
        \end{tabular}
        &
        \begin{tabular}{c}
        \\[-0.2cm]
        \textbf{chosen value} \\[0.2cm]
        \end{tabular} \\
        \hline
        
        \begin{tabular}{c}
        \\[-0.2cm]
        number of hidden layers
        \\[0.2cm]
        \end{tabular}
        &
        \begin{tabular}{c}
        \\[-0.2cm]
        $\{1,2 \}$
        \\[0.2cm]
        \end{tabular}
        &
        \begin{tabular}{c}
        \\[-0.2cm]
        $2$
        \\[0.2cm]
        \end{tabular} \\
        \hline
        
        \begin{tabular}{c}
        \\[-0.2cm]
        size of both hidden layers
        \\[0.2cm]
        \end{tabular}
        &
        \begin{tabular}{c}
        \\[-0.2cm]
        $\{25, 50, 100 \}$
        \\[0.2cm]
        \end{tabular}
        &
        \begin{tabular}{c}
        \\[-0.2cm]
        $50$
        \\[0.2cm]
        \end{tabular} \\
        \hline
        
        \begin{tabular}{c}
        \\[-0.2cm]
        batch size
        \\[0.2cm]
        \end{tabular}
        &
        \begin{tabular}{c}
        \\[-0.2cm]
        $\{50, 100 \}$
        \\[0.2cm]
        \end{tabular}
        &
        \begin{tabular}{c}
        \\[-0.2cm]
        $50$
        \\[0.2cm]
        \end{tabular} \\
        \hline
        
        \begin{tabular}{c}
        \\[-0.2cm]
        learning rate
        \\[0.2cm]
        \end{tabular}
        &
        \begin{tabular}{c}
        \\[-0.2cm]
        $\{0.001, 0.005, 0.01 \}$
        \\[0.2cm]
        \end{tabular}
        &
        \begin{tabular}{c}
        \\[-0.2cm]
        $0.005$
        \\[0.2cm]
        \end{tabular} \\
        \hline
        
        \begin{tabular}{c}
        \\[-0.2cm]
        weight decay
        \\[0.2cm]
        \end{tabular}
        &
        \begin{tabular}{c}
        \\[-0.2cm]
        $\{0.0001, 0.0005, 0.001 \}$
        \\[0.2cm]
        \end{tabular}
        &
        \begin{tabular}{c}
        \\[-0.2cm]
        $0.0005$
        \\[0.2cm]
        \end{tabular} \\
        \hline
        
        \begin{tabular}{c}
        \\[-0.2cm]
        patience
        \\[0.2cm]
        \end{tabular}
        &
        \begin{tabular}{c}
        \\[-0.2cm]
        $\{4, 8\}$
        \\[0.2cm]
        \end{tabular}
        &
        \begin{tabular}{c}
        \\[-0.2cm]
        $8$
        \\[0.2cm]
        \end{tabular} \\
        \hline
        
    \end{tabular}\\[4pt]
    \caption{Hyperparameter values for the case with 11-point feeedback.}
    \label{tab:hyperparam_11}
\end{table}

\section{Loss values for evaluation metrics related to RLHF}
\label{app:bin_loss_values}
\begin{figure}[htp]
    \centering
    \begin{subfigure}{0.3\textwidth}
        \centering
        \includegraphics[width=\textwidth]{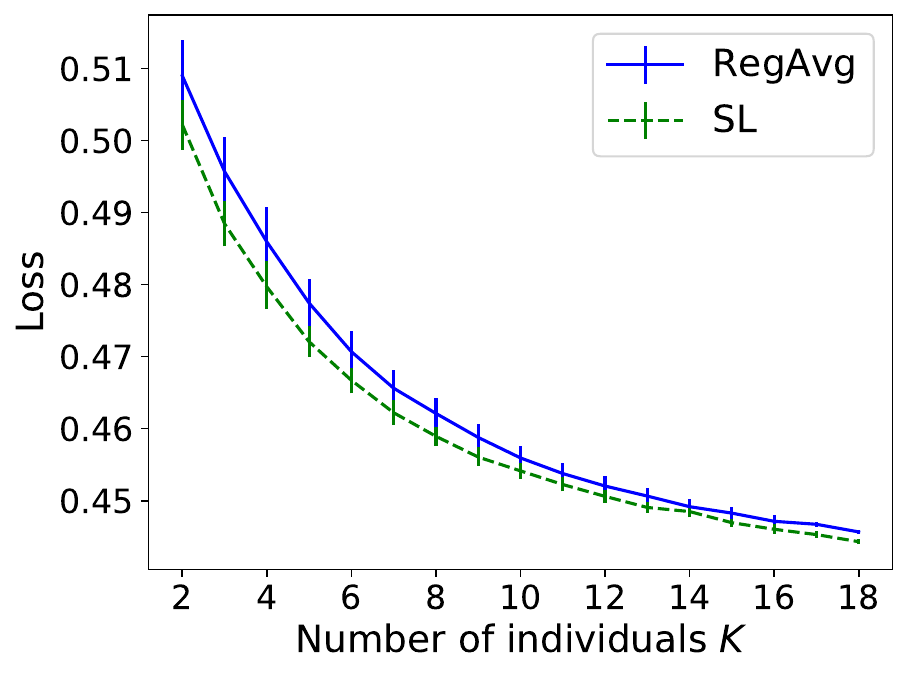}
        \caption{2-point feedback}
        \label{fig:loss_vs_K_2_ignore_neutral}
    \end{subfigure}
    \hfill
    \begin{subfigure}{0.3\textwidth}
        \centering
        \includegraphics[width=\textwidth]{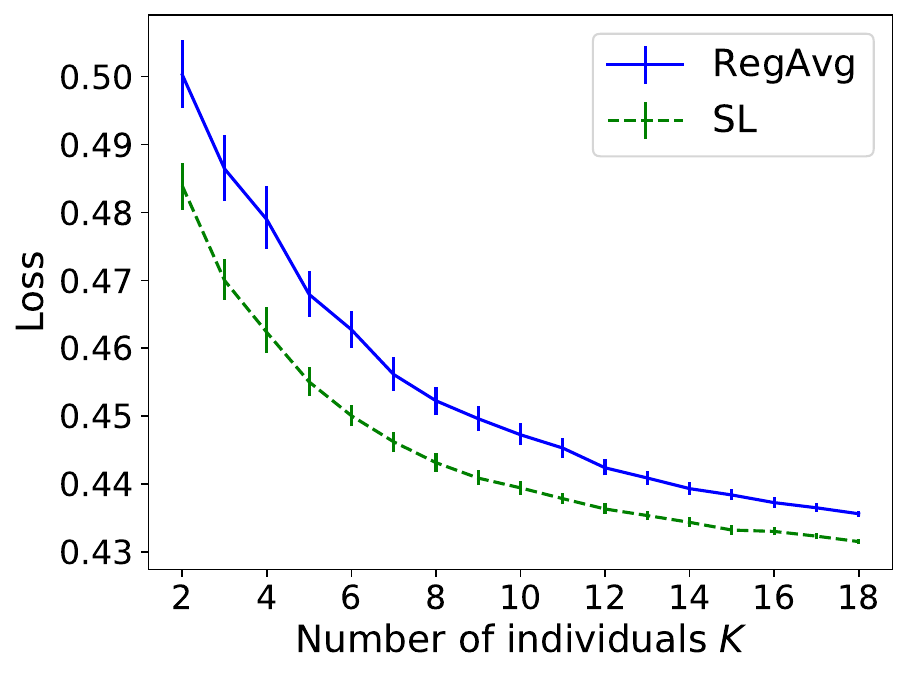}
        \caption{5-point feedback} 
        \label{fig:loss_vs_K_5_ignore_neutral}
    \end{subfigure} 
    \hfill
    \begin{subfigure}{0.3\textwidth}
        \centering
        \includegraphics[width=\textwidth]{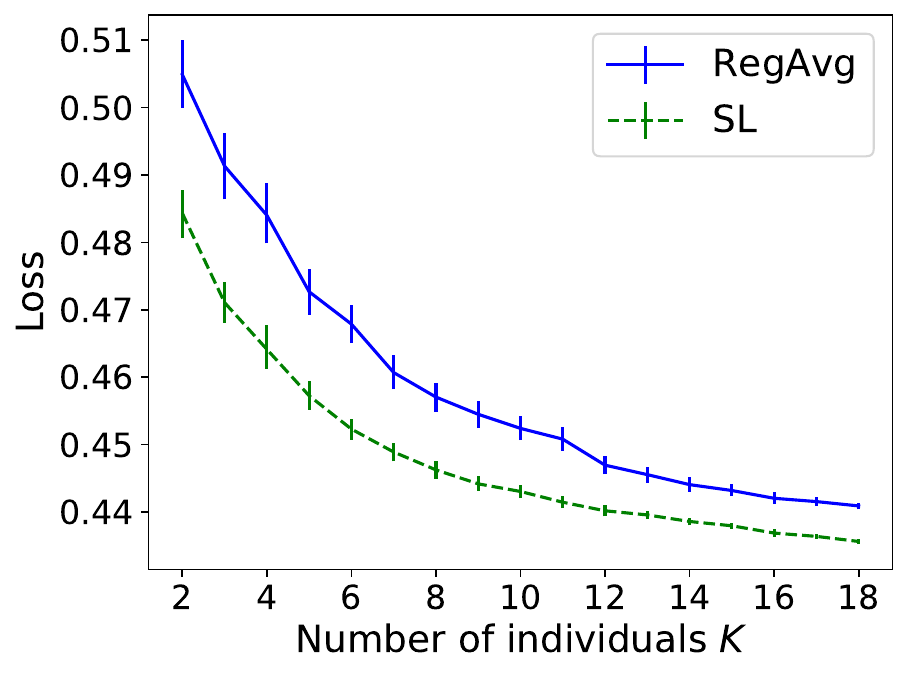}
        \caption{11-point feedback} 
        \label{fig:loss_vs_K_11_ignore_neutral}
    \end{subfigure}
    \caption{Losses of methods as a function of number of individuals for feedback of different granularities with loss function in Equation~\ref{eq:bin_loss_ignore_neutral}.}
    \label{fig:loss_vs_K_bin_losses_ignore_neutral}
\end{figure}

\begin{figure}[htp]
    \centering
    \begin{subfigure}{0.3\textwidth}
        \centering
        \includegraphics[width=\textwidth]{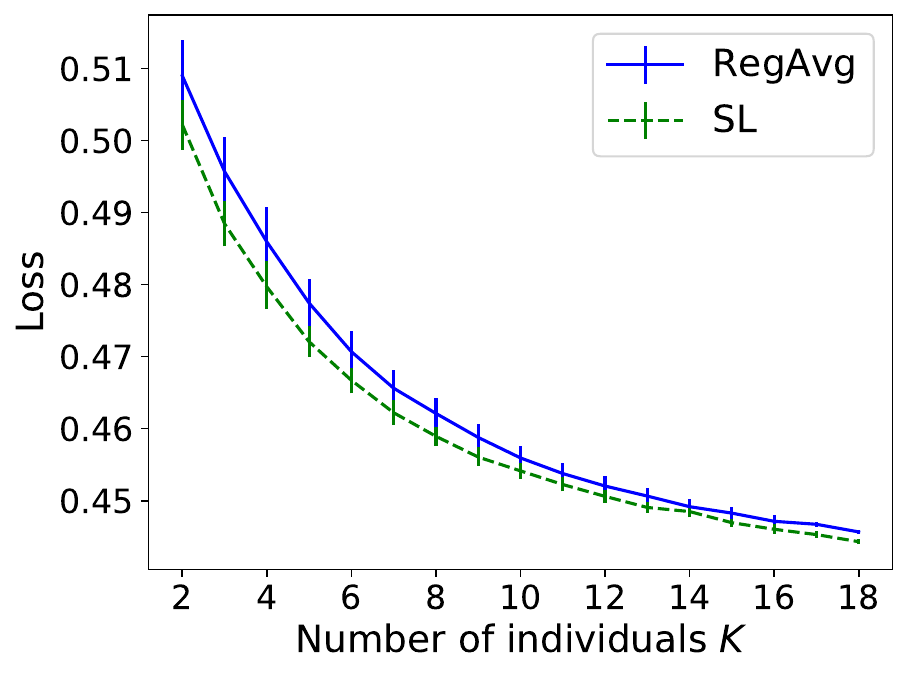}
        \caption{2-point feedback}
        \label{fig:loss_vs_K_2_keep_neutral}
    \end{subfigure}
    \hfill
    \begin{subfigure}{0.3\textwidth}
        \centering
        \includegraphics[width=\textwidth]{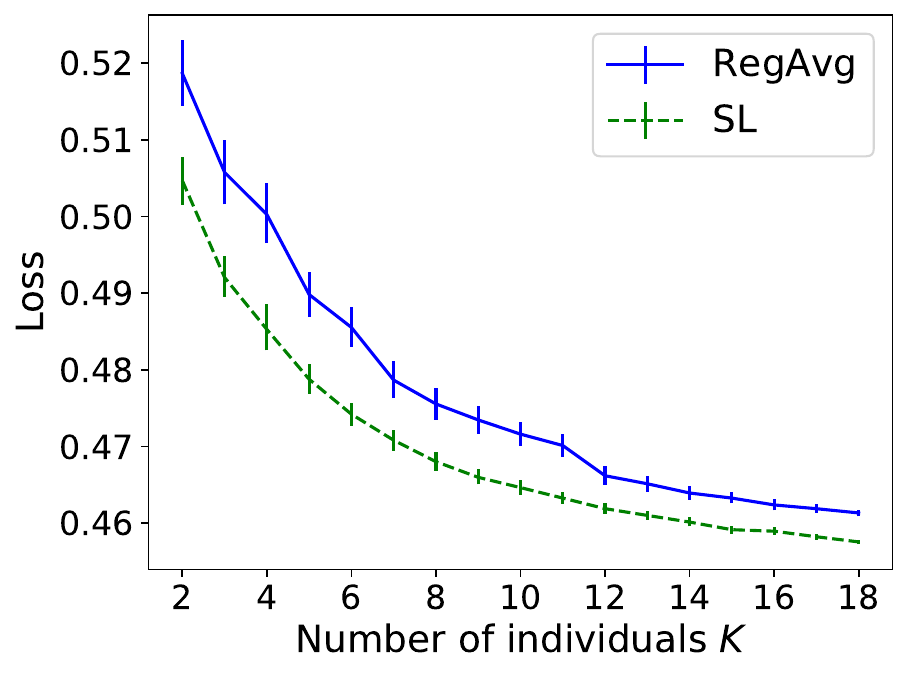}
        \caption{5-point feedback} 
        \label{fig:loss_vs_K_5_keep_neutral}
    \end{subfigure} 
    \hfill
    \begin{subfigure}{0.3\textwidth}
        \centering
        \includegraphics[width=\textwidth]{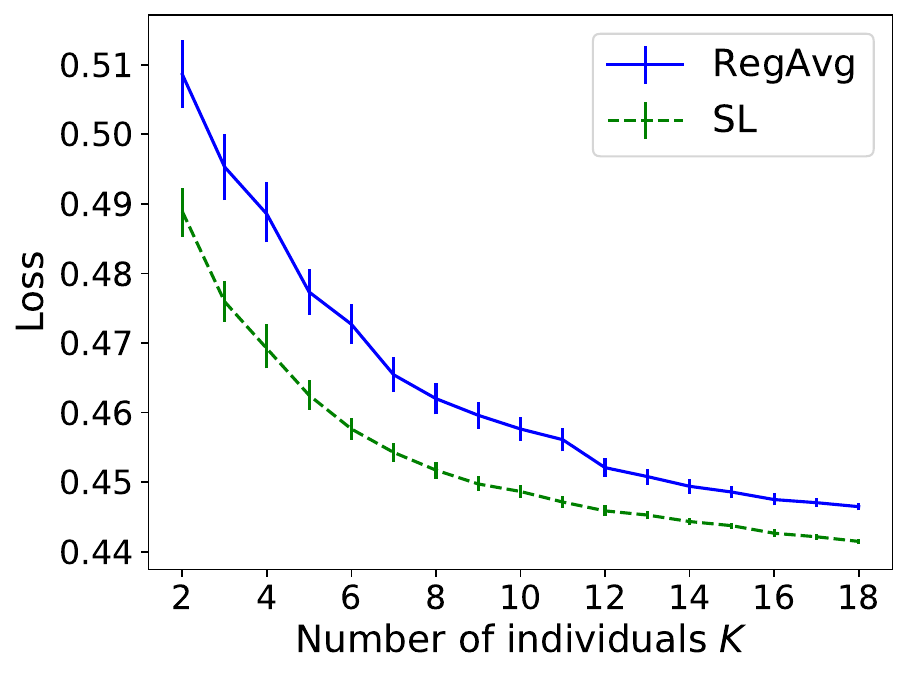}
        \caption{11-point feedback} 
        \label{fig:loss_vs_K_11_keep_neutral}
    \end{subfigure}
    \caption{Losses of methods as a function of number of individuals for feedback of different granularities with loss function in Equation~\ref{eq:bin_loss_keep_neutral}.}
    \label{fig:loss_vs_K_bin_losses_keep_neutral}
\end{figure}

In this section, we plot the average loss suffered by different methods for the evaluation metrics proposed in Section~\ref{sec:exp_details_bin_loss}. These plots are given in Figure~\ref{fig:loss_vs_K_bin_losses_ignore_neutral} and Figure~\ref{fig:loss_vs_K_bin_losses_keep_neutral}.

\end{document}